\documentclass[10pt,twocolumn,letterpaper]{article}

\usepackage{cvpr}              % To produce the CAMERA-READY version
\usepackage[dvipsnames]{xcolor}

\usepackage{graphicx}
\usepackage{amsmath}
\usepackage{amssymb}
\usepackage{booktabs}
\usepackage{microtype}
\usepackage{enumitem}
\usepackage{xcolor}
\usepackage[accsupp]{axessibility}

\makeatletter % keep these before authblk 
\robustify\@latex@warning@no@line
\makeatother 
\usepackage{authblk}
\makeatletter % keep these after authblk
\renewcommand\AB@affilsepx{~ \protect\Affilfont}

\robustify\@latex@warning@no@line
\makeatother

\usepackage{ulem}

\usepackage{comment}

\usepackage{amsmath}

\usepackage{diagbox}
\usepackage{multicol}
\usepackage{multirow}
\usepackage{float}

\newcommand{\jason}[1]{\textcolor{blue}{#1}}

\definecolor{cvprblue}{rgb}{0.21,0.49,0.74}
\usepackage[pagebackref,breaklinks,colorlinks,citecolor=cvprblue]{hyperref}

\usepackage[capitalize]{cleveref}
\creflabelformat{equation}{#2#1#3}
\crefname{section}{Sec.}{Secs.}
\Crefname{section}{Section}{Sections}
\Crefname{table}{Table}{Tables}
\crefname{table}{Tab.}{Tabs.}

\def\paperID{12019} % *** Enter the Paper ID here
\def\confName{CVPR}
\def\confYear{2024}

\begin{document}

%\title{Plausible Multi-view Object Insertion}
%\title{Generative Multi-view Scene Editing via Neural Radiance Fields using Near-to-Far Dataset Updates}
%\title{Diffusing Objects into Neural Radiance Fields for Multi-View Scene Editing}
%\title{View-Consistent Object Manipulation in Neural Radiance Fields\\using Near-to-Far Dataset Updates}
%\title{From Near to Far: Fusing Objects into Neural Radiance Fields\\ with Pose-Ordered Dataset Updates}

\title{Language-driven Object Fusion into Neural Radiance Fields\\ with Pose-Conditioned Dataset Updates}

%\title{Language-driven Neural Radiance Fields Editing through \\Pose-Conditioned Dataset Updates}

\author[1]{Ka Chun Shum}
\author[1]{Jaeyeon Kim}
\author[2,4]{Binh-Son Hua}
\author[3]{Duc Thanh Nguyen}
\author[1]{Sai-Kit Yeung}

\affil[1]{Hong Kong University of Science and Technology}
\affil[2]{VinAI}
\affil[3]{Deakin University}
\affil[4]{Trinity College Dublin}

\maketitle
\begin{comment}
\twocolumn[{
\begin{@twocolumnfalse}
\maketitle
\vspace{-0.2cm}
%\includegraphics[width=\linewidth,trim={0.1cm 4cm 0cm 0.1cm}]{images/teaser/teaser_dummy.pdf}
\includegraphics[width=\linewidth,trim={0.1cm 7cm 0cm 0.1cm}]{images/teaser/teaser_v2.pdf}
\captionof{figure}{\textbf{Object insertion}. We propose a language-driven method for consistent 3D object insertion in a background NeRF scene. Our method accomplishes plausible and diverse results that requires complicated geometry manipulation from the original background NeRFs. \textcolor{blue}{Todo: one more row to add.}}
\label{fig:teaser}
\vspace{0.8cm}
\end{@twocolumnfalse}
}]
\end{comment}

%%%%%%%%% ABSTRACT
\begin{abstract}
Neural radiance field (NeRF) is an emerging technique for 3D scene reconstruction and modeling. However, current NeRF-based methods are limited in the capabilities of adding or removing objects. This paper fills the aforementioned gap by proposing a new language-driven method for object manipulation in NeRFs through dataset updates. Specifically, to insert an object represented by a set of multi-view images into a background NeRF, we use a text-to-image diffusion model to blend the object into the given background across views. The generated images are then used to update the NeRF so that we can render view-consistent images of the object within the background. To ensure view consistency, we propose a dataset update strategy that prioritizes the radiance field training based on camera poses in a pose-ordered manner. We validate our method in two case studies: object insertion and object removal. Experimental results show that our method can generate photo-realistic results and achieves state-of-the-art performance in NeRF editing.
 
%with camera views close to the already-trained views prior to propagating the training to remaining views. We show that under the same dataset update strategy, we can easily adapt our method for object insertion using data from text-to-3D models as well as object removal. Experimental results show that our method can generate photorealistic images of the edited scenes, and outperforms state-of-the-art methods in 3D reconstruction and neural radiance field blending.

\end{abstract}

\begin{figure}[h]
  \centering
    \includegraphics[width=\columnwidth,trim={0cm 0cm 0cm 0cm}]{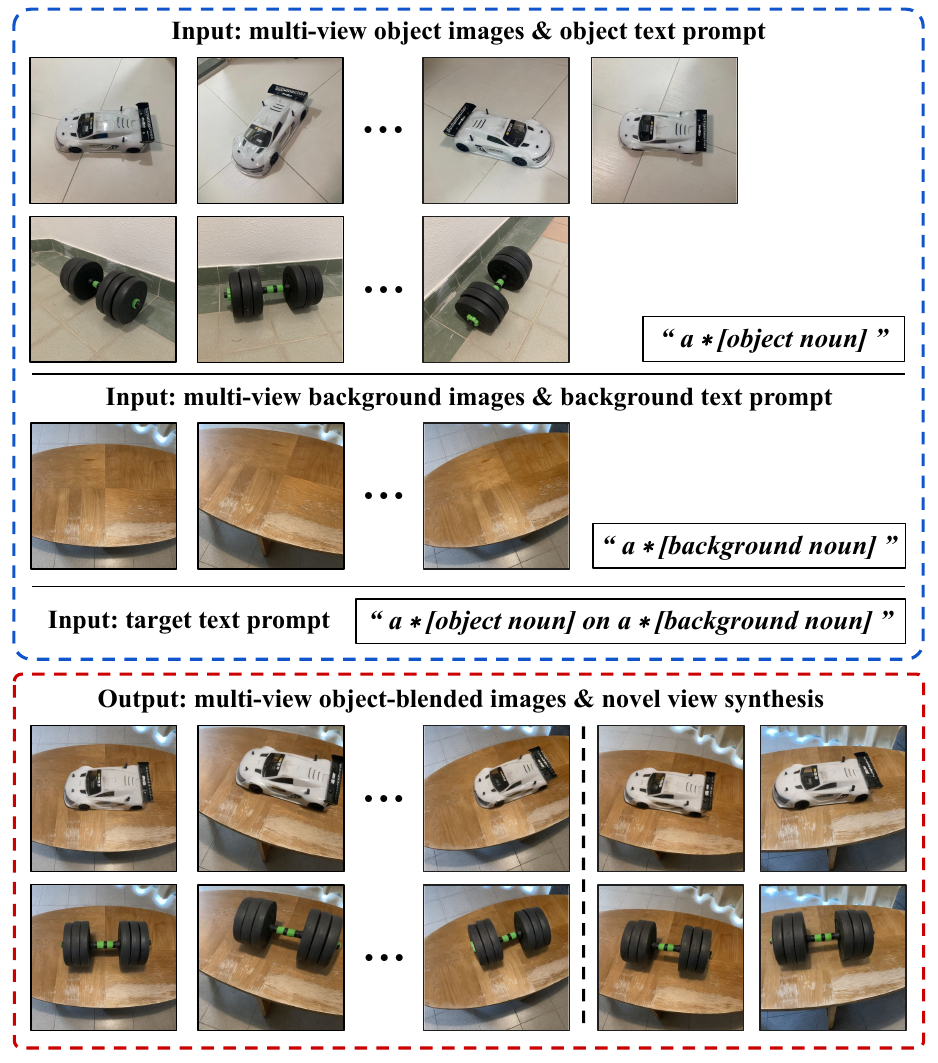}
    \vspace{-0.15in}
    \caption{\textbf{Object insertion}. We propose a language-driven method for view-consistent 3D object insertion into a background NeRF scene. Given an object defined by a set of multi-view images, our method generates plausible text-guided insertion results that require geometry manipulation of the original background NeRF.}
    \label{fig:teaser}
\end{figure}

%%%%%%%%% BODY TEXT
\section{Introduction}
Editing of 3D scenes by insertion or removal of objects has been a fundamental task in computer graphics and computer vision. The task has often been performed using traditional 3D scene authoring tools~\cite{blender2018,castellino2005computer}. For example, to insert an object into a 3D scene, traditional approach requires user to manually select the object and position it into the scene. This manual pipeline has been adopted in a wide range of applications, such as furniture arrangement in interior design~\cite{harper2012mastering} and asset creation in game development~\cite{haas2014history,unrealengine}. 

Recent advances in deep learning have opened new directions to scene modeling. Neural radiance field (NeRF)~\cite{mildenhall2021nerf} is a pioneer to render view-consistent photo-realistic images using neural networks. Generative models such as generative adversarial networks (GANs)~\cite{goodfellow2014generative} and diffusion models~\cite{ho2020denoising} learn to output photo-realistic images from unconstrained image collections. Recent text-guided diffusion models~\cite{rombach2022high,ramesh2022hierarchical,saharia2022photorealistic} have shown great promise in generating high-quality and diverse images from a single text prompt. 

%In this paper, we propose a new language-driven method for editing neural radiance fields and manipulating their objects. 

%\textcolor{blue}{as shown in~\cref{fig:teaser}, }we focus on the task of inserting a foreground object into a background radiance field. Our approach to this problem is to 

Such successes inspire us to revisit 3D scene editing via language-driven image synthesis techniques. Particularly, for object insertion (see~\cref{fig:teaser}), our method performs view-consistent edits by taking as input a set of multi-view object images, multi-view background images, and a target text prompt.
We utilize a text-to-image diffusion model to continuously synthesize multi-view images containing both a target object and a background during NeRF training. These synthesized images can be iteratively used to refine the NeRF of the background to learn geometric and appearance features of the object. This approach to refining a radiance field is also known as dataset update~\cite{haque2023instruct}.
%\jason{Do we need to cite Instruct-NeRF2NeRF here for the term \emph{dataset updates}? And mention in the next sentence it cannot make obvious geometry changes like add or remove objects, which is a challenge of the \emph{dataset updates} in Instruct-NeRF2NeRF.}
However, a particular challenge is that the refinement process may result in unstable NeRF training, degrading its rendering quality due to inconsistent views synthesized. To address this issue, we propose a \textit{pose-conditioned} dataset update strategy that gradually engages the target object into the background, beginning at a randomly selected pose (view), then views close to the already-used views before propagating to views further away. We observe that this pose-conditioned strategy significantly improves the NeRF learning, reducing rendering artifacts and maintaining view-consistent rendering. In summary, we make the following contributions in our work. 

% which can be captured, crawled from the Internet, or rendered through graphics engines~\cite{unrealengine,haas2014history}

\begin{itemize}
    \item We propose a new framework for object manipulation in NeRFs via text-to-image diffusion. Our work can create high-quality 3D scenes from simple inputs (text prompts and multi-view images). 
    It, therefore, has the potential for conveniently building a variant of 3D object libraries that is applicable for the scene editing task.
    %\sout{It, therefore, has the potential for building rich and large-scale 3D object libraries.}
    
    \item We propose a pose-conditioned dataset update strategy that stabilizes the fusion of objects into a background NeRF, enabling view-consistent rendering. Our method requires no priors on the geometry or texture of the objects, unlike existing works relying on accurate meshes~\cite{zhuang2023dreameditor,li2023focaldreamer,chen2023text2tex,huang2023nerf}, depth maps~\cite{weder2023removing,mirzaei2023spin}, fine masks~\cite{song2023blending,wang2023inpaintnerf360}, semantics~\cite{bao2023sine}, or lighting assumption~\cite{wu2023nerf,zhang2021nerfactor}.
    
    \item  We showcase our method in two case studies: object insertion and object removal via a user-friendly manner. Specifically, we use a 3D bounding box to define the location for object insertion/removal. Box-based location requires only a rough orientation, which is easy to be adjusted and visualized using real-time NeRF GUIs~\cite{tancik2023nerfstudio,muller2022instant}.
    
    \item We conduct extensive experiments to validate key techniques of our method and demonstrate its state-of-the-art performance in NeRF editing.
    %as well as an ablation study to analyze different factors of our proposed method. 
\end{itemize}

\section{Related work}

%Object manipulation has been a long-standing research problem in computer graphics and computer vision. Here, we limit our discussion to neural network-based methods with a focus on generative models and NeRF-based techniques. 

%\noindent\textbf{Image Manipulation.} 

\paragraph{Image synthesis.} 
Early attempts in deep learning-based image synthesis have utilized generative models such as GANs~\cite{goodfellow2014generative}. 
Several methods combine supervised losses with adversarial losses to learn class-conditional GANs~\cite{mirza2014conditional} and image-to-image translations~\cite{zhu2017unpaired,isola2017image,wang2018high}. StyleGAN~\cite{karras2019style} and its variants~\cite{karras2020analyzing,karras2020training} perform image editing directly on latent representations, but such editings are not intuitive and thus difficult to control the output.

%For example, RePaint~\cite{lugmayr2022repaint} and SmartBrush~\cite{xie2023smartbrush} inpaint images with masks by denoising. ControlNet~\cite{zhang2023adding} fine-tunes a twin diffusion model that accepts custom input images. Instruct-Pix2Pix~\cite{brooks2023instructpix2pix} and Imagic~\cite{kawar2023imagic} retrain on an edit-text-to-image dataset to allow text-guided editing. 

Recently, diffusion models~\cite{ho2020denoising} have made significant progress in image synthesis with high-quality data samples constructed from sophisticated forward and denoising diffusion steps. In addition, vision-language models such as CLIP~\cite{radford2021learning} have made text prompts an intuitive condition to generate and edit images. These developments have led to text-guided diffusion models~\cite{rombach2022high,ramesh2022hierarchical,saharia2022photorealistic} which offer excellent synthesis quality. Downstream applications then can be built by fine-tuning these text-guided models to fit with the application domains~\cite{lugmayr2022repaint,xie2023smartbrush,zhang2023adding,brooks2023instructpix2pix,kawar2023imagic,ruiz2023dreambooth}. Our method follows this streamline where we use a text-to-image diffusion model~\cite{ruiz2023dreambooth} to guide the fusion of objects into a background. %Rather than simple model inferences, we interactively provide extra conditional information to the model by controlling the noise schedule during inferences.

%\noindent\textbf{Neural 3D modeling.}

\paragraph{Neural 3D modeling.} Image-based 3D modeling methods use convolutional neural networks (CNNs)~\cite{krizhevsky2012imagenet} to learn 3D structures from multiple images~\cite{huang2018deepmvs,zhou2016view,tulsiani2018multi,eslami2018neural,liu2018geometry}. However, CNNs often struggle to deal with complex shapes, texture, and lighting captured in the images. Follow-up works integrate differentiable rendering and represent 3D structures as neural surfaces~\cite{michalkiewicz2019implicit} or shapes~\cite{chen2019learning}. 3D-aware GANs integrates volume rendering into their generators to synthesize novel views from a single image~\cite{chan2022efficient,or2022stylesdf,schwarz2022voxgraf,skorokhodov2022epigraf}. NeRFs~\cite{mildenhall2021nerf} apply the same neural rendering approach, but are optimized on a ray rendering loss on multi-view input images. There are methods addressing the limitations of NeRFs in various aspects such as visual artifacts~\cite{barron2021mip}, data complexity~\cite{martin2021nerf}, camera poses~\cite{yu2021pixelnerf}, and computational efficiency~\cite{zhang2020nerf++,muller2022instant}. 

%\noindent\textbf{NeRF editing.}

\begin{figure*}[h]
  \centering
    \includegraphics[width=\linewidth,trim={0cm 5.25cm 0.75cm 0cm}]{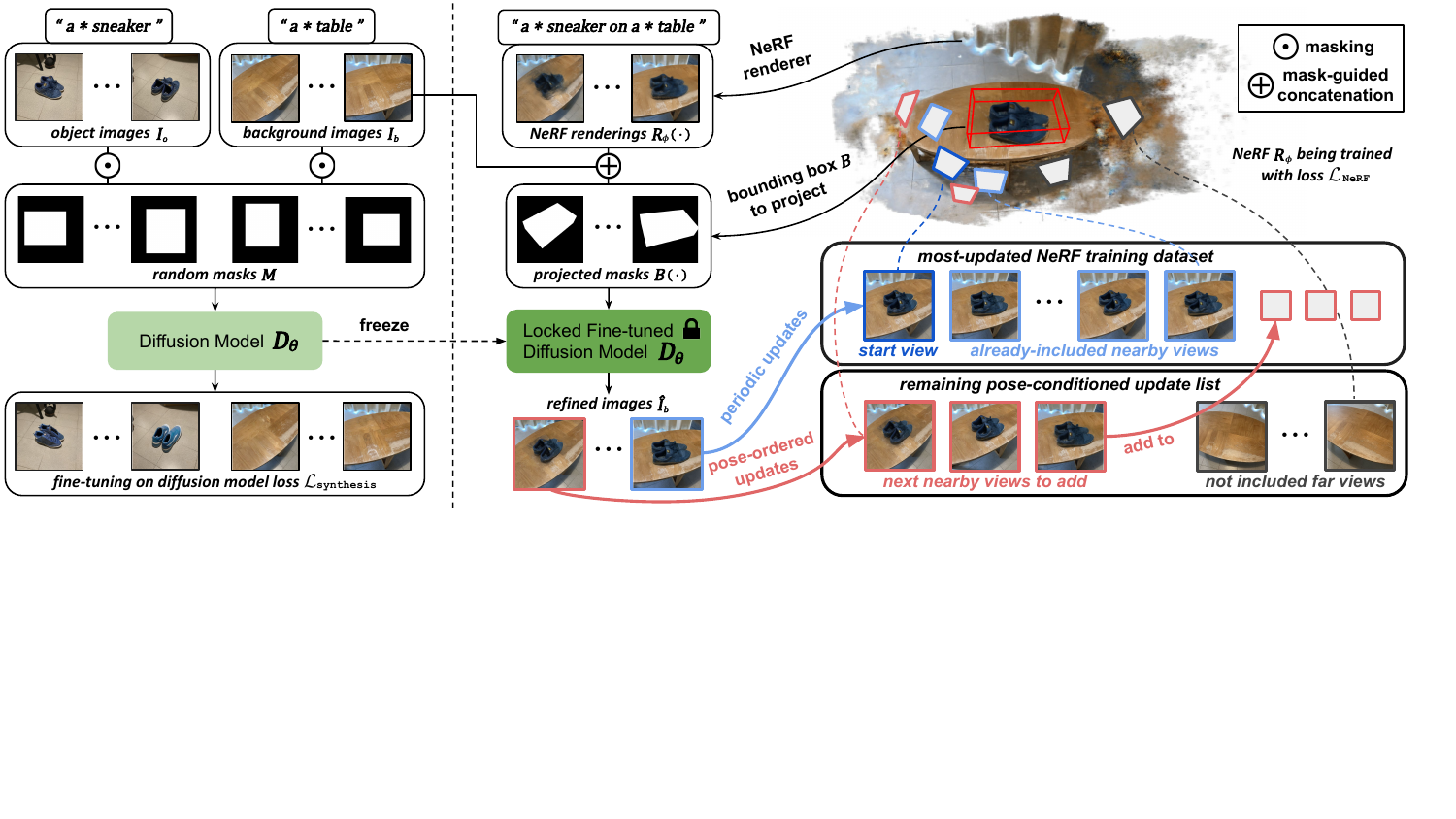}
    \vspace{-0.15in}
    \caption{\textbf{Overview of our pipeline}. We  customize and fine-tune a text-to-image diffusion model for view synthesis in an inpainting manner (left). We then apply the model to progressively fuse an object into background views to update a background NeRF (right). The process of view synthesis and NeRF updating is performed repeatedly. Views generated by the diffusion model are added to an on-going dataset to strengthen the NeRF. In return, the NeRF renders color hints for the diffusion model to create new views.}
    \label{fig:pipeline}
\end{figure*}

\paragraph{NeRF editing.} NeRF editing has often been carried out by parameter tuning~\cite{liu2021editing}, layer feature fusion~\cite{wang2022clip}, or deformable rays~\cite{yuan2022nerf}. An alternative to editing a NeRF is to directly amend the multi-view images used to learn it. For example, NeRF stylization methods~\cite{nguyen2022snerf,huang2022stylizednerf,wang2023nerf,pang2023local} freeze the geometry branch and optimize the color branch in a NeRF to stylize multi-view images. Several methods attempt to decompose an existing NeRF, which in turn holds color information and separates voxels considering multi-view masks~\cite{li20223ddesigner} or semantics~\cite{kim20233d,kobayashi2022decomposing}. NeRF inpainting fills simple unseen background geometry and colors with help of depth priors~\cite{mirzaei2023spin} or filtering inpainted multi-views~\cite{weder2023removing}. 

%BlendNeRF~\cite{kim20233d} automatically aligns poses and blends two NeRFs for human or animal face images. 

Creation of complex geometry and vivid colors for a NeRF is challenging due to higher-level requirements of consistency. DiscoScene~\cite{xu2023discoscene} fuses background and object NeRFs, thus is unable to condition customized contents. FocalDreamer~\cite{li2023focaldreamer} and DreamEditor~\cite{zhuang2023dreameditor} rely on fine meshes to function, disregarding the advantage of multi-view representation of NeRFs. Our method is perhaps most similar to Instruct-NeRF2NeRF~\cite{haque2023instruct} which shares a related data updating schema. However, Instruct-NeRF2NeRF~\cite{haque2023instruct} always generates geometry aligned with the original geometry (e.g., transfer a sneaker to a sneaker-shape apple), thus fails in most cases of object insertion or removals. Besides, it requires heavy retraining of a diffusion model on large-scale self-constructed datasets.

Recently, text-to-image diffusion has been applied to generate 3D contents. DreamFusion~\cite{poole2022dreamfusion} introduced a score distillation sampling (SDS) loss that progressively consolidates view information from a diffusion model into NeRFs. This loss has been applied with additional treatments to image resolution~\cite{lin2023magic3d}, conditional images~\cite{melas2023realfusion}, photo-realism~\cite{wang2023prolificdreamer}, or scene geometry~\cite{xu2023dream3d}. We also adopt the SDS loss but develop a novel training schedule to address challenges in multi-view object and background fusion for NeRF editing.

\section{Method}

\subsection{Overview} For the ease of presentation, we first describe our method for object insertion. We then extend it to object removal. An overview of our method for object insertion is illustrated in ~\cref{fig:pipeline}. Here we aim to insert an object into a background NeRF in two steps. In the first step, we synthesize training views for the NeRF with the target object embedded in the background. In the second step, the NeRF is updated with the synthesized views to learn the geometry and appearance of the object. % The training views in the first step can be created using generative image synthesis. 
%We expect the updated NeRF to be consistent with input views from the first step. 
%The advantage of this approach is that the radiance field can be trained progressively with updated data. 
There are two key challenges in this approach. First, background preserving is required in the synthesized views for the NeRF updating. Second, image synthesis may generate view-inconsistent images, causing artifacts in the resulting NeRF. 

To address the first challenge (i.e., object-blended image synthesis with background preserving), we leverage a state-of-the-art diffusion model to image synthesis, and opt to customize the model with text prompts for object blending (see~\cref{sec:fine-tuning}). %Specifically, we fine-tune a pre-trained text-to-image diffusion model \textcolor{blue}{to make it applicable to} generate images with the object blended into the background. 
We formulate this task as image inpainting where we place a binary mask on each background image to indicate the location where the object is inserted in, and adjust the text prompts with both the object and background. 
%\son{The binary mask can be generated from the 3D bounding box that indicates the position of the object in the background.}
%Here, we design our diffusion model based on the inpainting network of Stable Diffusion~\cite{rombach2022high}.
%\jason{DreamBooth is not doing inpainting, and learns on OBJ only. They generate a whole image. Learn on OBJ+BG and inference their combination, and under inpainting setting (we have different IO like random masks, etc, compared with original DreamBooth) should be one of our minor contribution.}
%We further build upon the interesting ability of DreamBooth~\cite{ruiz2023dreambooth} that personalizes the diffusion model to be able to generate images containing our object of interest and background. 
%\textcolor{blue}{We fine-tune thus customize the diffusion model where the object and background are specified by respective \textit{identifier}~\cite{ruiz2023dreambooth} tokens from text prompts. As a result, object-blended background images can be generated through the fine-tuned model by adjusting the text prompts.}

%This personalization is described by text prompts that contain special \textit{identifier} tokens for the object of interest and the background, respectively.
%Unfortunately, this approach only works well for single-image object insertion. Performing background for multiple camera views results in inconsistent results, making the training of the radiance field fail to converge to desirable quality.

To achieve view-consistent renderings, we propose a new strategy, namely \textit{pose-conditioned} dataset updates, to schedule the data used in the NeRF updating (see~\cref{sec:near_to_far_NeRF}). Our strategy is inspired by an important observation about the nature of NeRF: a view rendered by a NeRF maintains an extent of pose-aware color information from nearby already-used views, the nearer the more noticeable. Therefore, if we pass nearby renderings to the diffusion model with properly controlled noise, view-consistent results can be generated based on the learned color hints. 
%Reversely, the outputs from the fine-tuned diffusion model can be used to update the NeRF. As a consequence, the prior object knowledge from the diffusion model in 2D is gradually consolidated into the background NeRF in 3D.  
%Based on such observation, we design a novel data scheduler to our NeRF training as follows. We first optimize the NeRF by following the traditional training procedure on multi-view background images. We then progressively fuse the target object into this background NeRF by iteratively updating the dataset in a pose-ordered manner based on the distance from the object of interest to the camera position,
From such an observation, we design a novel data scheduler for our NeRF updating. We initially train the NeRF using regular method on a dataset of multi-view background images. We then progressively fuse the object into the NeRF by iteratively updating the dataset with object-blended background images in a pose-ordered manner, i.e., views are sorted in such a way that new views are acquired nearby already-used views. Our method can also be adapted to implement object removal (see~\cref{sec:object_removal}).

\subsection{Object-blended image synthesis}
\label{sec:fine-tuning}
%We aim to perform multi-view consistent object insertion to a given background via simple text prompts. 
We present our target object and background by a set of multi-view object images $\mathbf{I}_o$ and background images $\mathbf{I}_b$. Our goal is to build an image synthesis model that can blend the object (from $\mathbf{I}_o$) into the background (from $\mathbf{I}_b$) at custom locations specified by binary masks $M$. 

%Note that $\mathbf{I}^o$ and $\mathbf{I}^b$ can be real-world photographs or produced by image synthesis methods.
% $\hat{I} \Leftarrow (I^b, I^o, B)$

Let $D_\theta$ (with parameters $\theta$) be such an image synthesis model. Here we adopt a pre-trained Stable Diffusion~\cite{rombach2022high} to implement $D_\theta$. The model $D_\theta$ includes a denoising model $\epsilon_\theta$ that learns the noise component $\epsilon_\alpha \sim \mathcal{N}(\mathbf{0}, \mathbf{1})$ added in the diffusion process, where $\alpha \in [0,1]$ is a denoising strength. We customize $D_\theta$ with our background and object images in an inpainting fashion. Specifically, for each image $I \in \mathbf{I}_b \cup \mathbf{I}_o$, we make a masked-out (background-preserved) version $I \odot M$ where $M \sim \mathcal{U}$ is a random square binary mask whose coordinates are sampled from a uniform distribution $\mathcal{U}$ and $\odot$ is a pixel-wise product. The image $I$ is associated with a text prompt $p$ where we use ``$*$'' to indicate our object/background identifiers~\cite{ruiz2023dreambooth}, e.g., ``$*$ sneaker'' means that our target object is a sneaker and ``$*$ table'' is our target background. We fine-tune $D_\theta$ with the following loss:
\begin{equation}
    \mathcal{L}_\mathrm{synthesis}(\theta) = \mathbb{E}_{I, M \sim \mathcal{U}, p, \epsilon_\alpha \sim \mathcal{N}(\mathbf{0}, \mathbf{1})} \big[ \| \epsilon_\alpha - \epsilon_\theta (I) \|^2_2 \big]
  \label{eq:finetune_loss}
\end{equation}
%where $z$ is a latent representation generated by an encoder in the difussion model $D_\theta$.
%\jason{Should $\epsilon_\theta (p)$ be $\epsilon_\theta$ be enough already? As $\epsilon_\theta (p)$ appears only here and strictly $p$ is not the only internal input of $\epsilon_\theta$. In practice $\epsilon_\theta$ is a UNet accepts also the noisy image ($z_t$ in stable diffusion paper equation 2). We can keep it though if we want to obviously show $p$ before~\cref{eq:finetune}.}
%
%\son{Please check this loss function. Is the loss a diffusion loss? If so, $\epsilon_\theta$ should take an image as input for denoising?}
\cref{eq:finetune_loss} represents a general form for the fine-tuning process. Intermediate steps, e.g., input formation, latent encoding-decoding, are presented in our supplementary material.

%\begin{equation}
%    \hat{\mathcal{N}}(t) = D_\theta\left(\mathbf{I}^k + \mathcal{N}(t), M, \mathcal{N}(t), P^k\right),
%  \label{eq:finetune}
%\end{equation}
%where $p$ is a text prompt containing identifier tokens~\cite{ruiz2023dreambooth}, for either the object or background. 

%To fine-tune $D_\theta$, we use custom text-image pairs that include a user-defined token to identify the background and object, e.g., we use ``$*$'' as an adjective to modify a sneaker noun, meaning that an ``$*$ sneaker'' is our target object. Such text prompt is then paired with a few images of the same object, e.g., photos of the same sneaker, for fine-tuning the diffusion model. The fine-tuned text-to-image model can then generate our background and object instance with unseen camera poses.

%To conveniently refer to the target background and object with text prompts, we use a similar identifier word in DreamBooth~\cite{ruiz2023dreambooth} to construct our background prompt $P^b$ and object prompt $P^o$. Typically, such prompts could be of the form "a [identifier] [class noun]" (e.g., "a sks sneaker"). We use two different identifiers for background and object, respectively. The identifiers then act as an activation key later for the fine-tuned diffusion model to generate content about the target background or object. 

%from unseen camera poses 
We fine-tune the diffusion model $D_\theta$ with all the images in $\mathbf{I}_b$ and $\mathbf{I}_o$ for $n_{bg}$ and $n_{obj}$ times. After fine-tuning, we can use $D_\theta$ to synthesize images $\hat{I}_b$ that blend the object presented in $\mathbf{I}_o$ into the background images $I_b \in \mathbf{I}_b$ via a combined text prompt $\Tilde{p}$, e.g., ``a $*$ sneaker on a $*$ table'',
\begin{align}
    D_\theta\left(I_b \odot M, \Tilde{p}, \alpha \right) \rightarrow \hat{I}_b
  \label{eq:finetune}
\end{align}
where $\alpha$ is set manually to control how much object content in the masked area to be inpainted by $D_\theta$ (we demonstrate how we set $\alpha$ for different purposes in~\cref{sec:near_to_far_NeRF}).
%\jason{, e.g., force $\alpha=1$ (maximum strength) to generate the object presented in $\mathbf{I}_o$ in the masked area randomly from pure noise.}
%\son{What is the value range of $\alpha$? Is this alpha similar to alpha in $\epsilon_\alpha$?}

%as no object clues are available in the masked areas.

%\textcolor{blue}{\sout{We fine-tune $D_\theta$ with all the images in $\mathbf{I}^b$ and $\mathbf{I}^o$ for $n_{bg}$ and $n_{obj}$ times, respectively, and use the fine-tuned model to generate view-consistent images $\hat{\mathbf{I}}$ containing both the object and background for the subsequent NeFR updating.}}

%for the subsequent NeRF refinement stage in~\cref{sec:near_to_far_NeRF}.

%The fine-tuned text-to-image model is then used to generate background and object images from unseen camera poses.

%%%%%%%%%%%%%%%%%%%%%%%%%%%%%%%%%%%%%%%%%%%%%%%%%%%%%%%%%%%%%
\subsection{Pose-conditioned dataset updates}
\label{sec:near_to_far_NeRF}

%Our method is built on Instant-NGP~\cite{muller2022instant}, an off-the-shelf method for standard NeRF training. 
%We propose to train the NeRF on a dataset that is updated progressively.

Let $R_\phi$ (with parameters $\phi$) be our NeRF, which is initially trained with the multi-view background images in $\mathbf{I}_b$ using an existing approach~\cite{muller2022instant}. We also retrieve a set of camera poses $\{\pi(I_b)\}$ ($4\times4$ matrices) for the images $I_b \in \mathbf{I}_b$ using the pose estimation method COLMAP~\cite{schoenberger2016sfm,schoenberger2016mvs}. 
%\son{I suggest use $\pi$ instead of $P$ for camera pose.}
%We denote $P(I_b) \in \mathbf{P}(\mathbf{I}_b)$ as the camera pose for an image $I_b \in \mathbf{I}_b$.

We propose to update the NeRF $R_\phi$ progressively with object-blended background images generated by the fine-tuned diffusion model $D_\theta$. Our idea is to build a progressive multi-view and pose-conditioned object-blended image dataset $\hat{\mathbf{I}}_b$ from $\mathbf{I}_b$ to be used to update $R_\phi$. We initialize $\hat{\mathbf{I}}_b^{(0)}=\emptyset$ and $\mathbf{I}^{(0)}_b = \mathbf{I}_b$. We start with a random background image $I_b \in \mathbf{I}_b$. Let $B$ be a 3D bounding box, and $B(I_b)$ be the projection mask of $B$ onto $I_b$ using the camera pose $\pi(I_b)$. We generate an object-blended image $\hat{I}_b$ as,
\begin{align}
    %\hat{I}_{noisy}({T_0}) & = I^b({T_0}) + (\mathcal{N}(t=1) \oplus B(T_0)), \\
    D_\theta \left(I_b \odot B(I_b), \Tilde{p}, \alpha\right) \rightarrow \hat{I}_b
  \label{eq:first_view_generation}
\end{align}
where we set $\alpha=1$ (maximum strength) to blend the object into the masked area by using only the knowledge that $D_\theta$ learns from $\Tilde{p}$ (during the fine-tuning), and colors from the background image $I_b$ as there is no existing color clue in the masked area $I_b \odot B(I_b)$. 

We update $\hat{\mathbf{I}}^{(1)}_b = \hat{\mathbf{I}}^{(0)}_b \cup \{\hat{I}_b\}$ and $\mathbf{I}^{(1)}_b =\mathbf{I}^{(0)}_b \setminus \{I_b\}$. We then update $R_\phi$ using $\hat{\mathbf{I}}^{(1)}_b$, and keep doing so by adding images into $\hat{\mathbf{I}}_b$ sequentially. In particular, let $\hat{\mathbf{I}}_b^{(n-1)}$ be the dataset at the $(n-1)$-th step. The next image $I_b^{(n)} \in \mathbf{I}_b$ for step $n$ is selected so as it is closest (in terms of the camera poses) to already-used images,
\begin{align}
    I_b^{(n)} = \mathop{\arg\min}_{I_{b,k} \in \mathbf{I}^{(n)}_b} \mathop{\min}_{I_{b,j} \in \mathbf{I}^{(n-1)}_b} \| \pi_T(I_{b,k}) - \pi_T(I_{b,j}) \|^2
    \label{eq:distance_calculation}
\end{align}
where $\pi_T$ is the translation component of the pose $\pi$. 
This view selection reflects the standard multi-view reconstruction pipeline where the multi-view data are collected with a smoothly connected, inward-surrounding camera trajectory. This requirement is also fulfilled by most NeRF datasets.

Given a background image $I_b^{(n)}$ in~\cref{eq:distance_calculation}, to utilize the nearby color hints learned by $R_\phi$, we create a background-preserved foreground-rendered image $\tilde{I}_b^{(n)}$ as,
\begin{align}
    \tilde{I}_b^{(n)} = (I_b^{(n)} \odot B(I_b^{(n)})) \oplus (R_\phi(\pi(I_b^{(n)})) \odot \bar{B}(I_b^{(n)}))
    \label{eq:mixed_images}
\end{align}
%\son{Check brackets.}
where $R_\phi(\pi(I_b^{(n)}))$ is the rendering result of $R_\phi$ in the pose $\pi(I_b^{(n)})$ (for the background image $I_b^{(n)}$), $\bar{B}$ is the complement of $B$ (as we want to maintain the foreground rendered by $R_\phi$), and $\oplus$ is a pixel-wise addition.
%\son{This fusion is best illustrated with a small inline diagram.}

We then generate a view-consistent image $\hat{I}_b^{(n)}$ by applying $D_\theta$ to $\tilde{I}_b^{(n)}$ defined in~\cref{eq:mixed_images} as,
\begin{align}
    D_\theta\left(\tilde{I}_b^{(n)} \odot B(I_b^{(n)}), \Tilde{p}, \alpha \right) \rightarrow \hat{I}_b^{(n)}
  \label{eq:near_view_generation}
\end{align}
where we set $\alpha$ to a low value to allow $D_\theta$ to utilize color hints from previous nearby views, provided from the rendering results $R_\phi(\pi(I_b^{(n)})$. We empirically found that $\alpha \in [0.3,0.4]$ gives best view-consistent rendering.

%\jason{The noisy version of $\tilde{I}(T_i)$ includes useful object information to be used by diffusion model (which is not shown in~\cref{eq:near_view_generation}). By controlling $\alpha$, we can decide how much object content in the masked area to keep and repair by $D_\theta$. This is an important observation (and technique) of our method as it is related to why we concatenate in~\cref{eq:mixed_images}(we want rough obj info from NeRF rendering), why we do pose-conditioned updates(to make sure this rough obj info from NeRF is view consistent), and why we do customization(to make sure given the rough obj info diffusion model can still repair it by noising-and-denoising).}

Again, we update $\hat{\mathbf{I}}_b^{(n)}$ with $\hat{I}_b^{(n)}$, and update $R_\phi$ accordingly by minimizing the loss:
\begin{equation}
    \mathcal{L}_\mathrm{NeRF}(\phi) = \mathbb{E}_{\hat{I}_b \in \hat{\mathbf{I}}_b^{(n)}} \big[ \| R_\phi(\pi(\hat{I}_b)) - \hat{I}_b \|^2 \big].
  \label{eq:nerf_loss}
\end{equation}

During the updating process, we include $n_{near}$ views into the ongoing dataset for every $n_{new}$ NeRF updating steps. In addition, we periodically replace each old view by a new one using~\cref{eq:mixed_images,eq:near_view_generation} for every $n_{old}$ NeRF updating steps. The updating procedure is completed once all the background images in $\mathbf{I}_b$ have been processed, i.e., $\mathbf{I}_b^{(n)}=\emptyset$. 
%\son{Give an example of $n_{near}, n_{new}, n_{old}$.}

%the last view is included and the NeRF is updated for another $n_{new}$ steps.

%\subsection{Integrating Text-to-3D Object}
%\label{sec:text_to_3D}

%Our method can also support text-driven object insertion from synthetic 3D models. In particular, given a target object prompt, we use a text-to-3D method such as DreamFusion~\cite{poole2022dreamfusion} to generate a NeRF that represents a 3D object, which we can perform rendering to generate multi-view images for the object $I^o$, which can be used as input to our pipeline. From our experience, having a number of fixed camera trajectories inward-surrounding the object is sufficient to construct a valid $I^o$ for most objects.

\subsection{Adapting to object removal}
\label{sec:object_removal}

Our framework can be adapted to object removal. In particular, we also fine-tune the diffusion model $D_\theta$ as in~\cref{sec:fine-tuning}. However, we do not fine-tune $D_\theta$ with object images as we want to remove objects.
Instead, we first individually inpaint all background images on projection masks using background prompts without identifier (e.g., ``a table'').
%\textcolor{blue}{\sout{Instead, we inpaint multi-view background images on projection masks using background prompts without identifier (e.g., ``a table'').}}
These inpainted images are then treated as \textit{pseudo ground-truth} and used to customize $D_\theta$ with the background text prompts containing identifiers (e.g., ``a $*$ table''). The pseudo ground-truth backgrounds are visually pleasing but still remain cross-view inconsistencies. To circumvent this issue, we perform NeRF updating using dataset updates as in~\cref{sec:near_to_far_NeRF}. However, only background prompts are used during the NeRF updating. We found that this joint 2D-3D interaction gradually transforms disruptive inpainted image regions into view-consistent background images. We show the necessity of the pseudo ground-truth and the NeRF updating in making view-consistent object removal in our experiments.

\section{Experiments}
\subsection{Datasets}
For object insertion, we propose a dataset comprising multi-view images of 8 backgrounds and 10 objects. The number of images for each background and object ranges from 60 to 100 and 40 to 80, respectively. We collected the images using an iPhone camera and resized the images to $512\times512$ resolution for the ease of training. Except for our self-captured data, we also test on synthetically rendered multi-view images~\cite{mildenhall2021nerf,poole2022dreamfusion}.

For object removal, we run experiments on the commonly used inpainting datasets from Mip-NeRF-360~\cite{barron2022mip} and IBRNet~\cite{wang2021ibrnet}. We centrally cropped images in those datasets to make $512\times512$ images to fit with our pipeline.

\subsection{Baselines}
%Our method composes information from multiple images into a consistent 3D. 
%To the best of our knowledge, we are the first to address the challenging view-consistent scene editing problem given purely the multi-view background and object images. 
For comparisons, we select SOTA baselines from different branches of work that can perform object insertion. To ensure the fairness, we barely modify the baselines, only when necessary to fit them with our settings. 

\noindent\textbf{Traditional 3D editing pipeline.} This pipeline creates a 3D model from multi-view images. Scene editing is then performed directly on the 3D model. To simulate this traditional pipeline, we use COLMAP~\cite{schoenberger2016sfm,schoenberger2016mvs} to reconstruct the mesh for the background and object, and then manually crop and place the object mesh into the background mesh. 

%To ensure minimal changes, we do not make additional manual rendering of texture and lighting information.

% baseline: point cloud + point cloud --> manual placement
% problems: lighting; boundary; geometry.

% Traditional image inpaiting fits well object removal~\cite{weder2023removing,mirzaei2023spin,wang2023inpaintnerf360}. 

\noindent\textbf{Image inpainting.} We adopt the inpainting variant of Stable Diffusion~\cite{rombach2022high} as a baseline. However, since image inpaiting treats each view independently, for a fair comparison, we perform single-view inpainting on each background image.

% baseline: stable diffusion inpainting (2023)
% problems: serious object inconsistency.

\begin{comment}
\noindent\textbf{Conditional 3D-aware GAN.}
3D-aware GAN is well-studied for generating images from different camera poses facing a same 3D content~\cite{chan2022efficient}. The conditional 3D-aware GAN is able to base the 3D on the given images. Although current methods do not accept multiple images, we still compare a SOTA conditional 3D-aware GAN baseline, 3D generation on ImageNet (hereinafter called 3DImageNet)~\cite{skorokhodov20233d}, which accepts a single image as condition. We use the same starting view mentioned in~\cref{eq:first_view_generation} to feed the baseline, which is a minimum change to the baseline to fit our task.
% baseline: 3D generation on ImageNet (ICLR 2023)
% problems: background and object mismatch; serious artifacts for too-faraway camera poses.
\end{comment}

%A NeRF can use knowledge from external models such as CNNs~\cite{yu2021pixelnerf}, vision transformers~\cite{lin2023vision}, and diffusion models~\cite{chen2023single}. 

\noindent\textbf{Single-image-to-3D NeRF.} NeRF-based view synthesis partially fulfills scene editing. Here we select Zero123~\cite{liu2023zero}, a SOTA that uses a distillation prior~\cite{poole2022dreamfusion} from a 2D diffusion model to synthesize novel views for comparison. 

%Similarly, we use the same starting view in~\cref{eq:first_view_generation} as the single image.

% baseline: Zero-1-to-3: Zero-shot One Image to 3D Object (2023)
% problems: background and object mismatch; serious artifacts for too-faraway camera poses; Fail to deal with complicated background.

%\noindent\textbf{NeRF blending.} 
%Given multi-view images of the object and background in our input setting, training two corresponding NeRFs and then blending them into one is possible for scene editing. We do not consider naive and manual blending as it is similar to the traditional computer graphics pipeline which we already compared. Instead, as a baseline, we use BlendNeRF~\cite{kim20233d} to combine two NeRFs automatically by specifying two masked areas in two key-frames in the two NeRFs.

% baseline: 3D-aware Blending with Generative NeRFs (ICCV 2023) (its GitHub says code release in Aug)
% problems: The automaticity fails to generalize on unseen image domain; The blending fails if two to-be-combined NeRFs are of different content domains.

\noindent\textbf{Instruct-NeRF2NeRF}~\cite{haque2023instruct}. This work also applies dataset updates for NeRF training. However, it fails to add/remove objects with noticeable non-uniform appearance. To test this method, we re-format text prompts, e.g., we replace ``a $*$ sneaker on a $*$ table'' by ``add/make a sneaker on a table''.

%We follow its default setting except for adapting text prompts to required format. For example, the text prompt is changed from ``a $*$ sneaker on a $*$ table'' to ``add a sneaker on a table''.

% baseline: Instruct-NeRF2NeRF: Editing 3D Scenes with Instructions (2023)
% problems: the geometry of the original NeRF will retain, thus fail to add object/remove objects.

%\noindent\textbf{Other methods.} DreamEditor~\cite{zhuang2023dreameditor} utilizes DreamBooth~\cite{ruiz2023dreambooth} for fine-grained editing, but fails to construct objects with unseen geometry. BlendNeRF~\cite{kim20233d} combines object and background NeRFs, but in limited image domains and thus it is not applicable to our diverse scene data. 

%Given multi-view object and background images as input, training two corresponding NeRFs and then blending them into one is possible for scene editing. We do not consider naive and manual blending, as it is similar to the traditional computer graphics pipeline, which we already compared. 

\subsection{Implementation details}
%Below configuration works for most of our editing cases. 

We fine-tune the diffusion model $D_\theta$ with object images ($n_{obj} = 5,000$ times) more than background images ($n_{bg} = 500$ times) as we found the model needs extra training to learn objects with complex geometry and texture. For NeRF training, we empirically found $n_{near}$ = 3, $n_{new}$ = 500, and $n_{old}$ = 10 balance well the quality and efficiency. 

%We fix the noise strength as mentioned in~\cref{eq:first_view_generation,eq:near_view_generation}.

We run all experiments on a Nvidia RTX 3090 GPU. Diffusion fine-tuning takes around 30 minutes. NeRF updating speed relies on the number of background images while each backpropagation takes 0.5 seconds. Inference speed of the diffusion model is 8 seconds/image. Our NeRF training costs around 2.7 times more than the standard NeRF training.

\subsection{Qualitative results}
\begin{figure*}[t]
    \hspace{0.6mm}sneaker on table\hspace{0.1mm}
    \includegraphics[width=0.08\paperwidth]{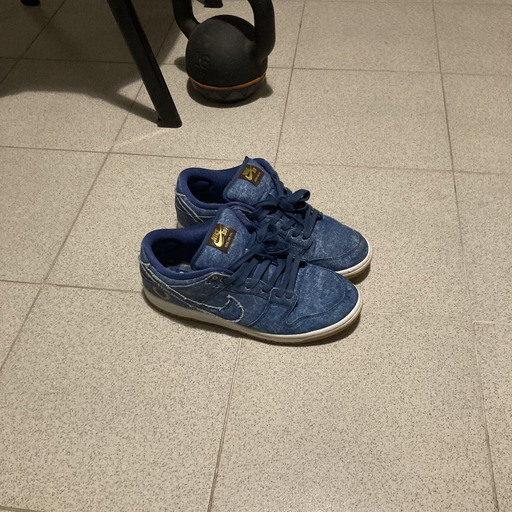}
    \hspace{1mm}backpack on wall\hspace{0.1mm}
    \includegraphics[width=0.08\paperwidth]{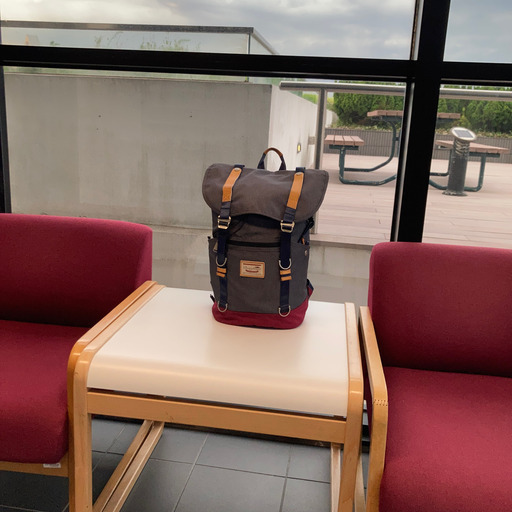}
    \hspace{1.5mm}table set in room\hspace{0.1mm}
    \includegraphics[width=0.08\paperwidth]{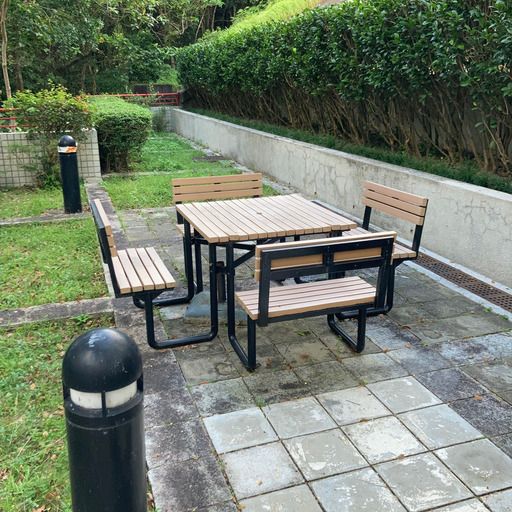}
    \hspace{7.0mm}car on road\hspace{1.0mm}
    \includegraphics[width=0.08\paperwidth]{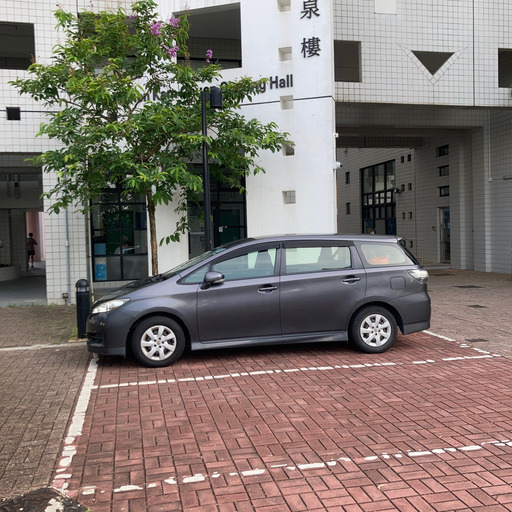}
   \\[0.8mm]
    \subfloat[Input]{\begin{minipage}[c]{\textwidth}
        \includegraphics[width=0.115\textwidth]{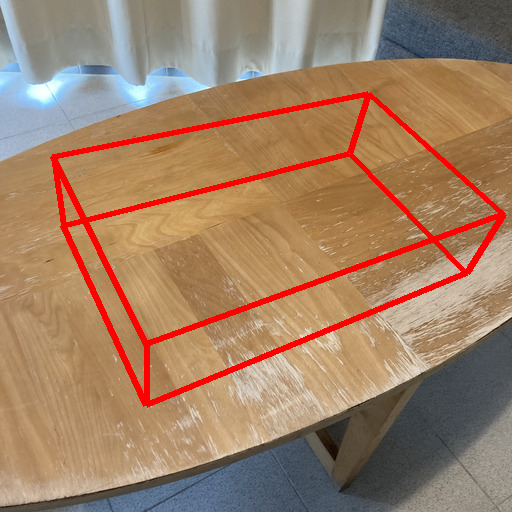}
        \hspace{0.0005\textwidth}
        \includegraphics[width=0.115\textwidth]{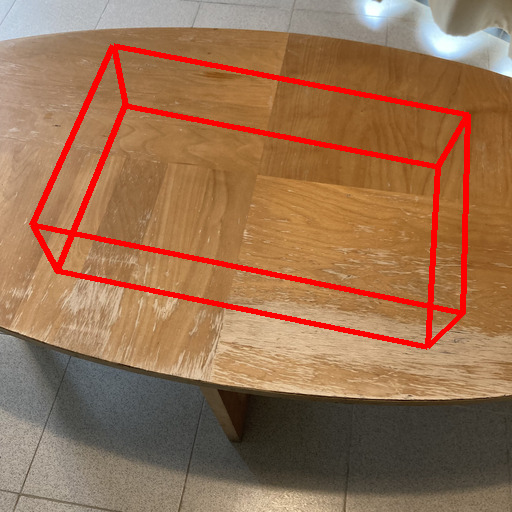}
        \hfill
        \includegraphics[width=0.115\textwidth]{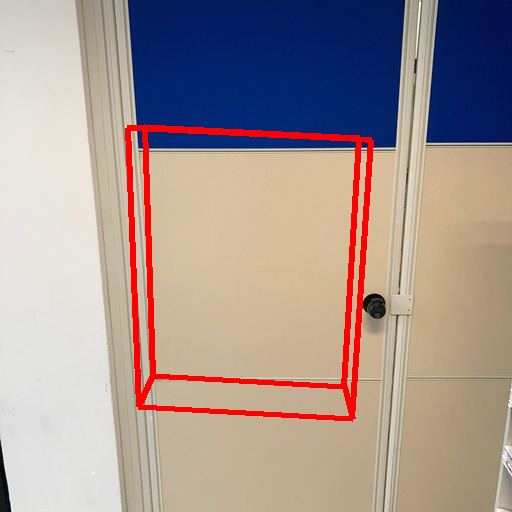}
        \hspace{0.0005\textwidth}
        \includegraphics[width=0.115\textwidth]{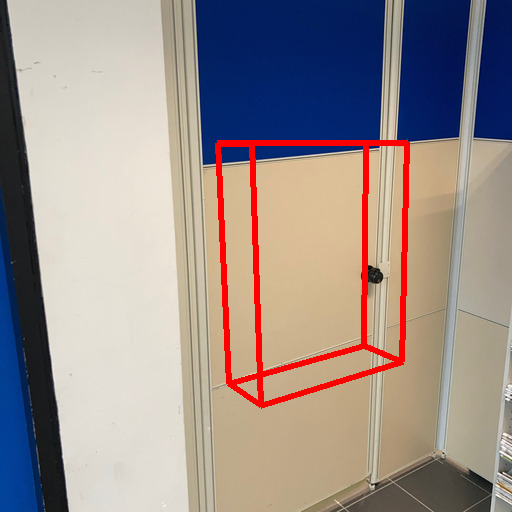}
        \hfill
        \includegraphics[width=0.115\textwidth]{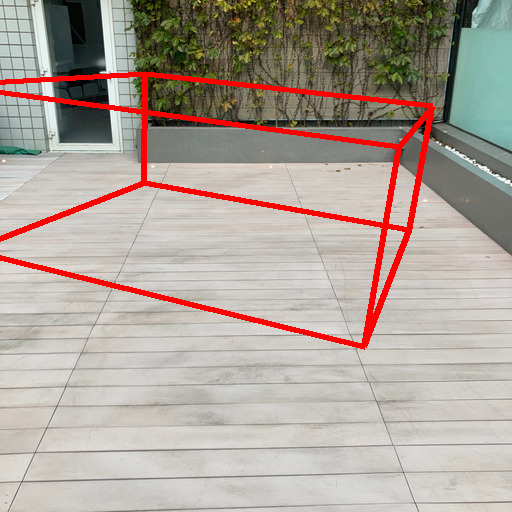}
        \hspace{0.0005\textwidth}
        \includegraphics[width=0.115\textwidth]{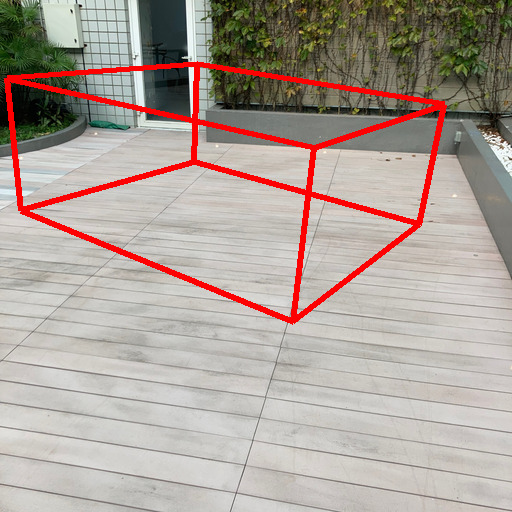}
        \hfill
        \includegraphics[width=0.115\textwidth]{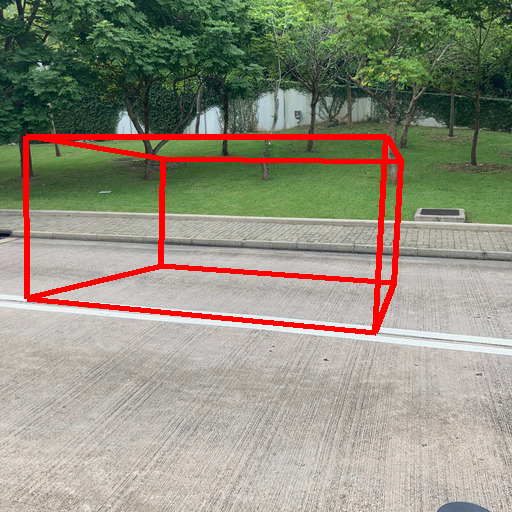}
        \hspace{0.0005\textwidth}
        \includegraphics[width=0.115\textwidth]{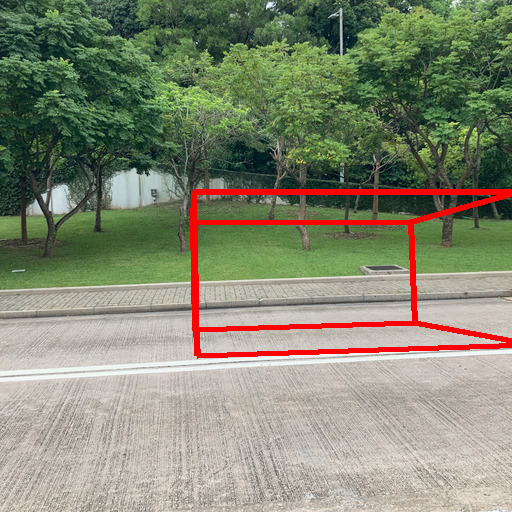}
    \end{minipage}}
    \smallskip
    
    \subfloat[Traditional 3D Pipeline (COLMAP mesh reconstruction~\cite{schoenberger2016sfm,schoenberger2016mvs} + manual placement)]{\begin{minipage}[c]{\textwidth}
        \includegraphics[width=0.115\textwidth]{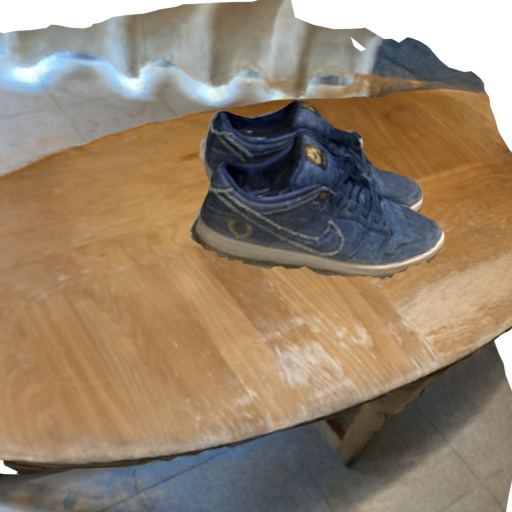}
        \hspace{0.0005\textwidth}
        \includegraphics[width=0.115\textwidth]{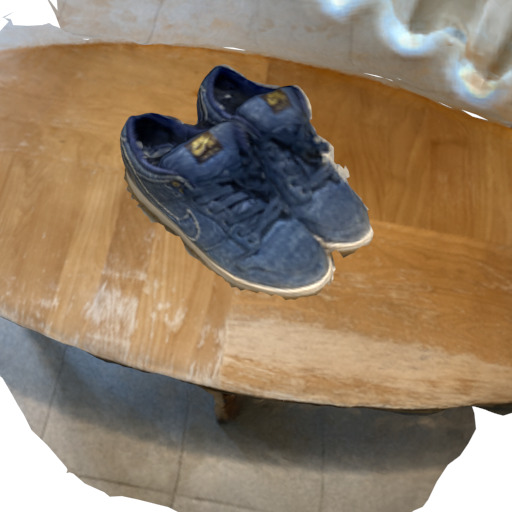}
        \hfill
        \includegraphics[width=0.115\textwidth]{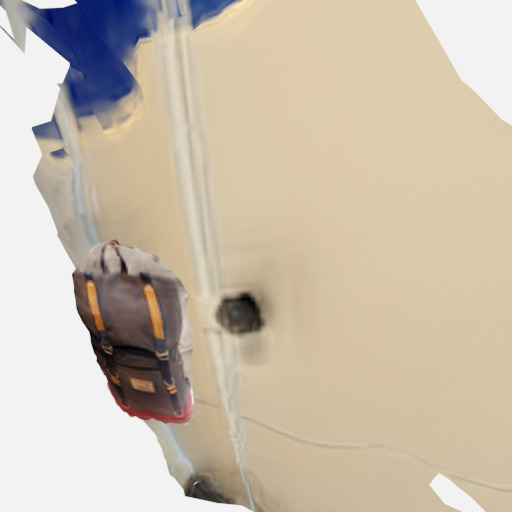}
        \hspace{0.0005\textwidth}
        \includegraphics[width=0.115\textwidth]{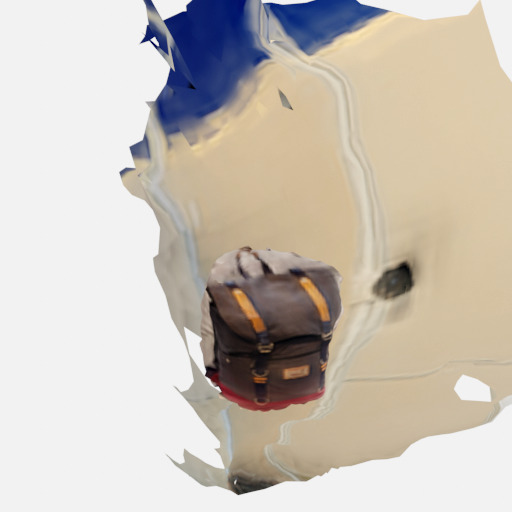}
        \hfill
        \includegraphics[width=0.115\textwidth]{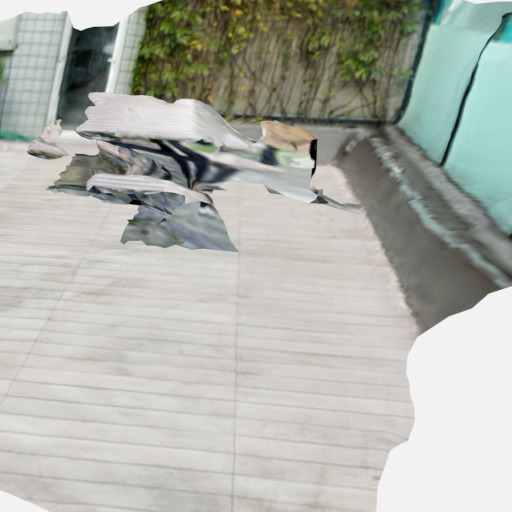}
        \hspace{0.0005\textwidth}
        \includegraphics[width=0.115\textwidth]{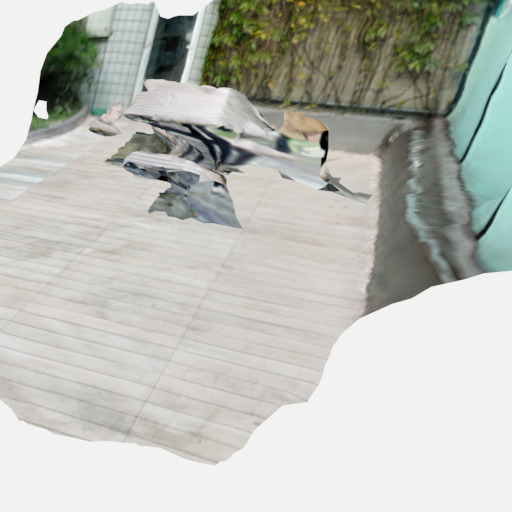}
        \hfill
        \includegraphics[width=0.115\textwidth]{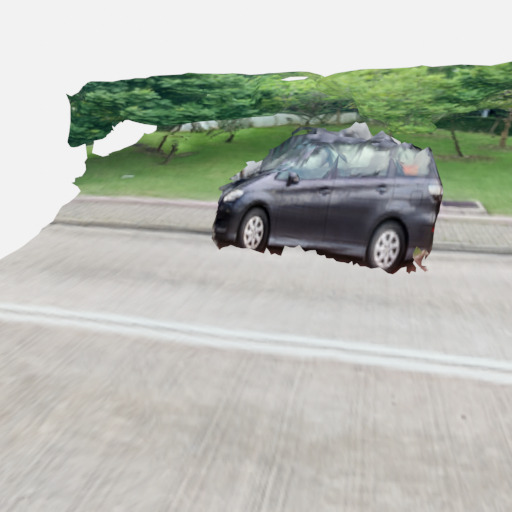}
        \hspace{0.0005\textwidth}
        \includegraphics[width=0.115\textwidth]{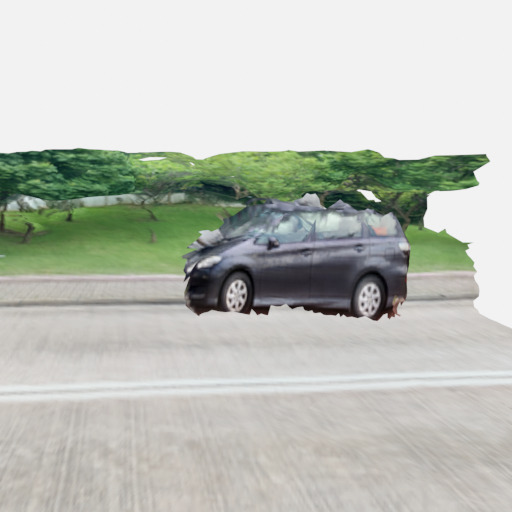}
    \end{minipage}}
    \smallskip

    \subfloat[Image Inpainting (Stable Diffusion Inpainting~\cite{rombach2022high})]{\begin{minipage}[c]{\textwidth}
        \includegraphics[width=0.115\textwidth]{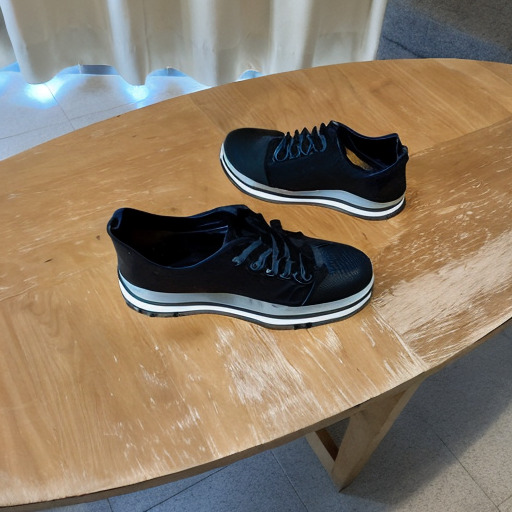}
        \hspace{0.0005\textwidth}
        \includegraphics[width=0.115\textwidth]{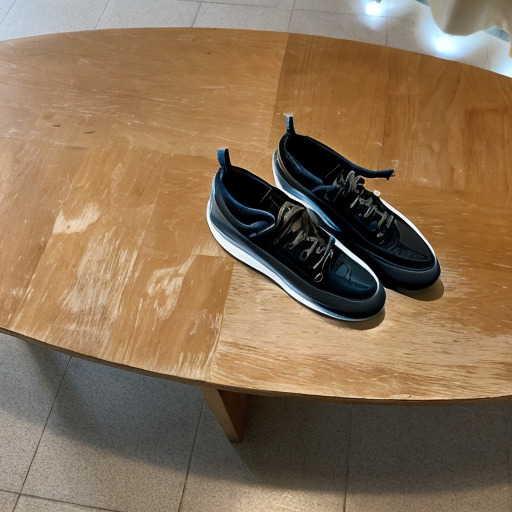}
        \hfill
        \includegraphics[width=0.115\textwidth]{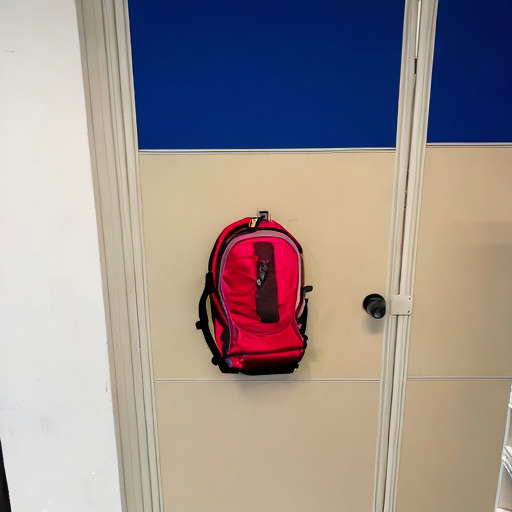}
        \hspace{0.0005\textwidth}
        \includegraphics[width=0.115\textwidth]{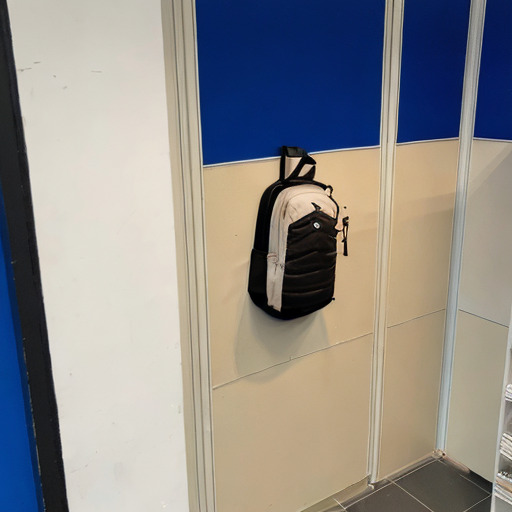}
        \hfill
        \includegraphics[width=0.115\textwidth]{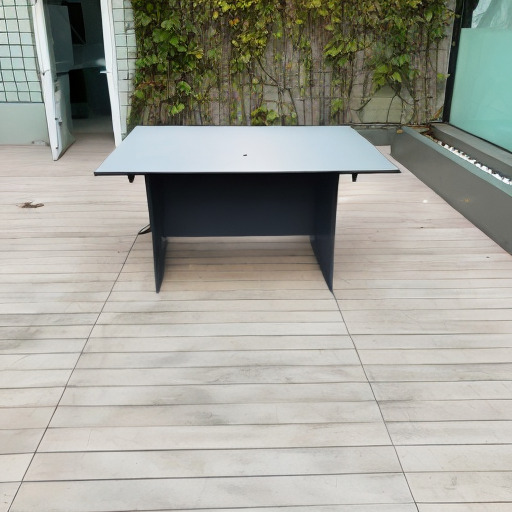}
        \hspace{0.0005\textwidth}
        \includegraphics[width=0.115\textwidth]{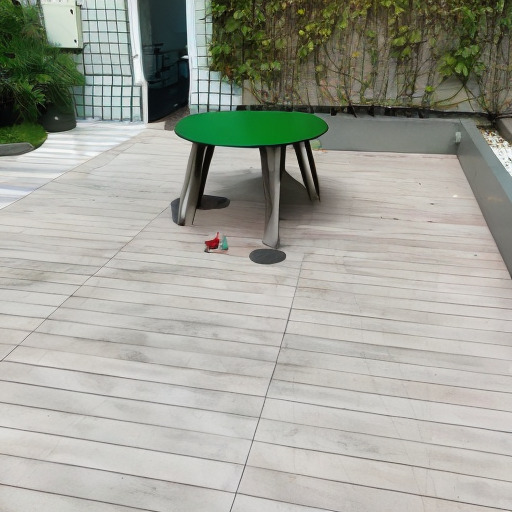}
        \hfill
        \includegraphics[width=0.115\textwidth]{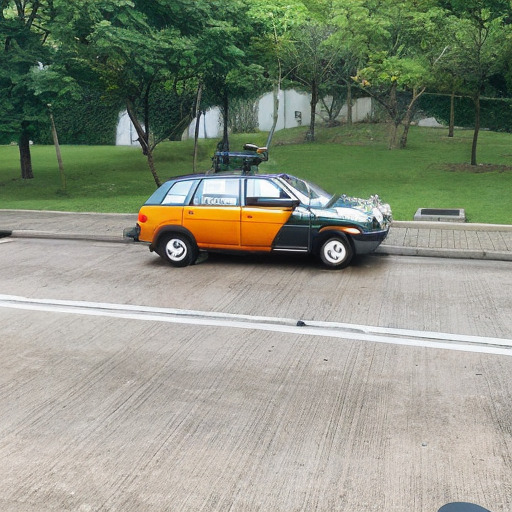}
        \hspace{0.0005\textwidth}
        \includegraphics[width=0.115\textwidth]{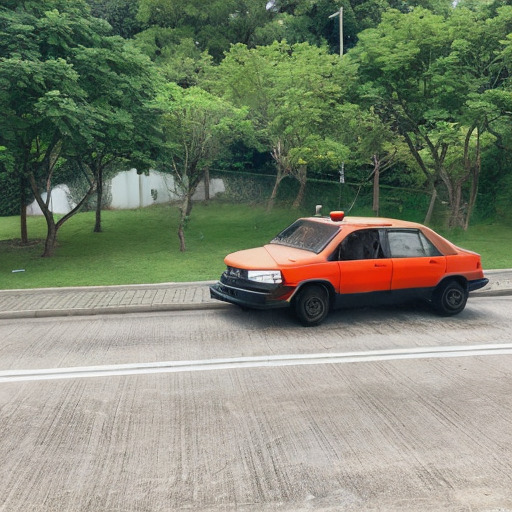}
    \end{minipage}}
    \smallskip
    
    %\subfloat[3D generation on ImageNet~\cite{skorokhodov20233d}]{\begin{minipage}[c]{\textwidth}
    %    \includegraphics[width=0.115\textwidth]{images/dummy_square.png}
    %    \hspace{0.0005\textwidth}
    %    \includegraphics[width=0.115\textwidth]{images/dummy_square.png}
    %    \hfill
    %    \includegraphics[width=0.115\textwidth]{images/dummy_square.png}
    %    \hspace{0.0005\textwidth}
    %    \includegraphics[width=0.115\textwidth]{images/dummy_square.png}
    %    \hfill
    %    \includegraphics[width=0.115\textwidth]{images/dummy_square.png}
    %    \hspace{0.0005\textwidth}
    %    \includegraphics[width=0.115\textwidth]{images/dummy_square.png}
    %    \hfill
    %    \includegraphics[width=0.115\textwidth]{images/dummy_square.png}
    %    \hspace{0.0005\textwidth}
    %    \includegraphics[width=0.115\textwidth]{images/dummy_square.png}
    %\end{minipage}}
    %\smallskip

    \subfloat[Single-image-to-3D NeRF (Zero123~\cite{liu2023zero})]{\begin{minipage}[c]{\textwidth}
        \includegraphics[width=0.115\textwidth]{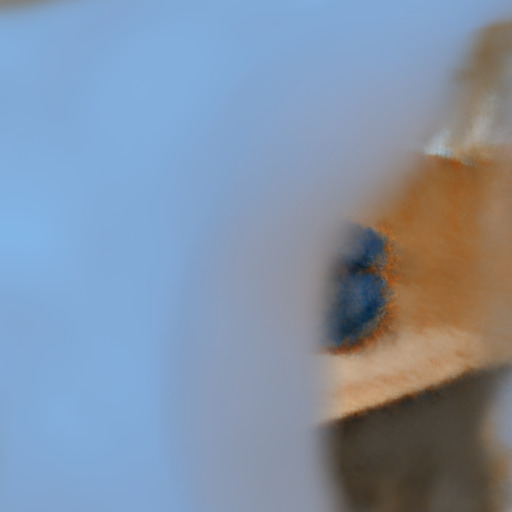}
        \hspace{0.0005\textwidth}
        \includegraphics[width=0.115\textwidth]{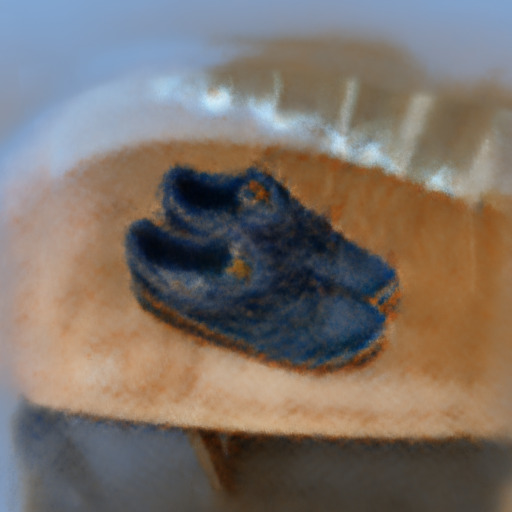}
        \hfill
        \includegraphics[width=0.115\textwidth]{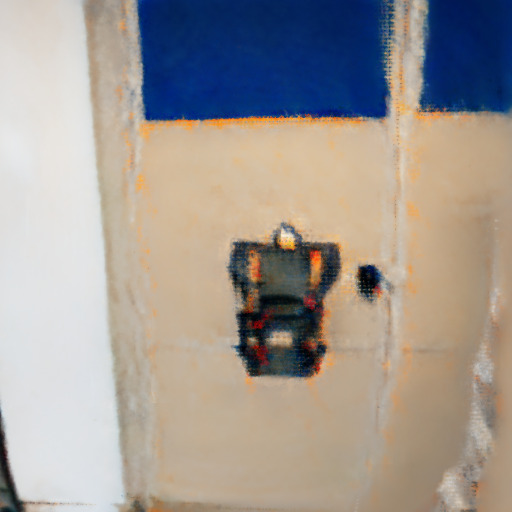}
        \hspace{0.0005\textwidth}
        \includegraphics[width=0.115\textwidth]{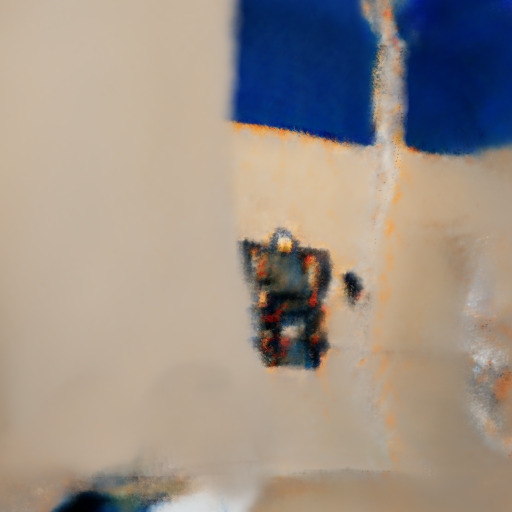}
        \hfill
        \includegraphics[width=0.115\textwidth]{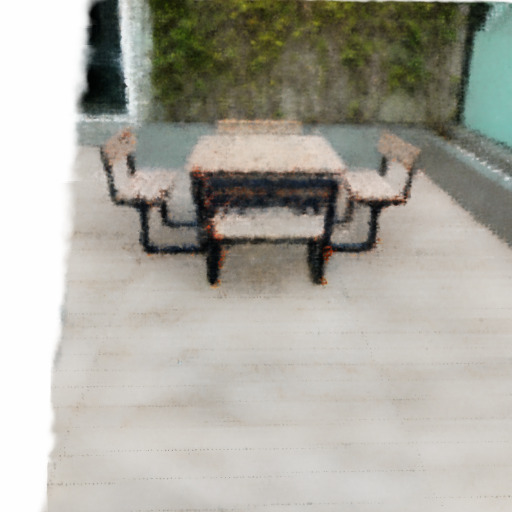}
        \hspace{0.0005\textwidth}
        \includegraphics[width=0.115\textwidth]{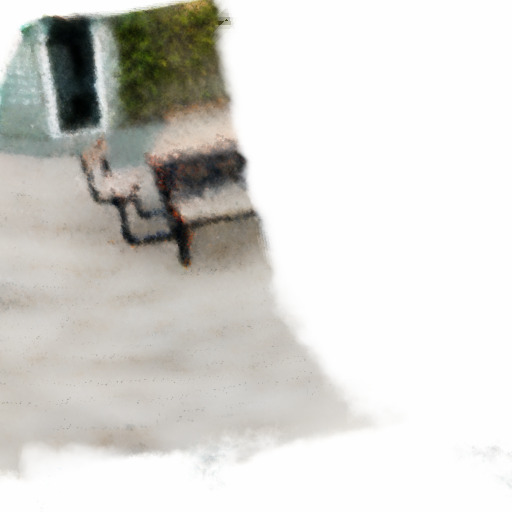}
        \hfill
        \includegraphics[width=0.115\textwidth]{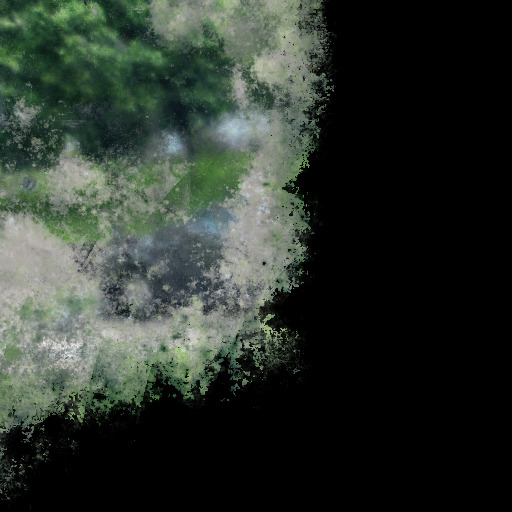}
        \hspace{0.0005\textwidth}
        \includegraphics[width=0.115\textwidth]{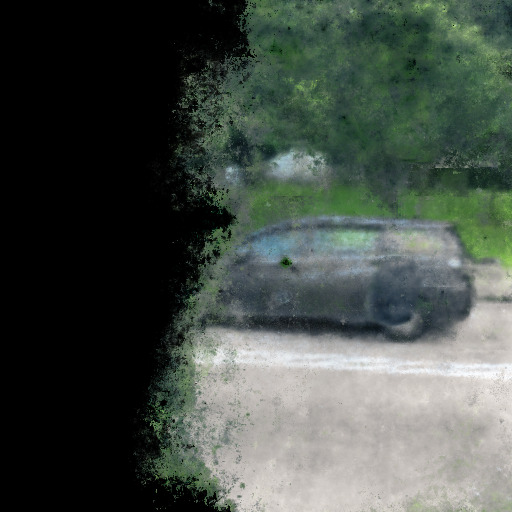}
    \end{minipage}}
    \smallskip

    %\subfloat[NeRF Blending (BlendNeRF~\cite{kim20233d})]{\begin{minipage}[c]{\textwidth}
    %    \includegraphics[width=0.115\textwidth]{images/dummy_square.png}
    %    \hspace{0.0005\textwidth}
    %    \includegraphics[width=0.115\textwidth]{images/dummy_square.png}
    %    \hfill
    %    \includegraphics[width=0.115\textwidth]{images/dummy_square.png}
    %    \hspace{0.0005\textwidth}
    %    \includegraphics[width=0.115\textwidth]{images/dummy_square.png}
    %    \hfill
    %    \includegraphics[width=0.115\textwidth]{images/dummy_square.png}
    %    \hspace{0.0005\textwidth}
    %    \includegraphics[width=0.115\textwidth]{images/dummy_square.png}
    %    \hfill
    %    \includegraphics[width=0.115\textwidth]{images/dummy_square.png}
    %    \hspace{0.0005\textwidth}
    %    \includegraphics[width=0.115\textwidth]{images/dummy_square.png}
    %\end{minipage}}
    %\smallskip

    \subfloat[Instruct-NeRF2NeRF~\cite{haque2023instruct}]{\begin{minipage}[c]{\textwidth}
        \includegraphics[width=0.115\textwidth]{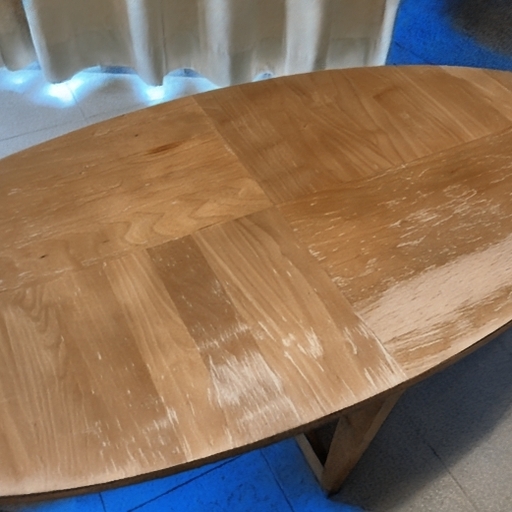}
        \hspace{0.0005\textwidth}
        \includegraphics[width=0.115\textwidth]{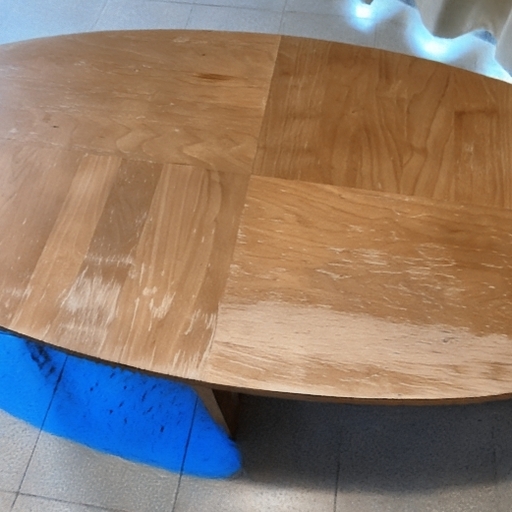}
        \hfill
        \includegraphics[width=0.115\textwidth]{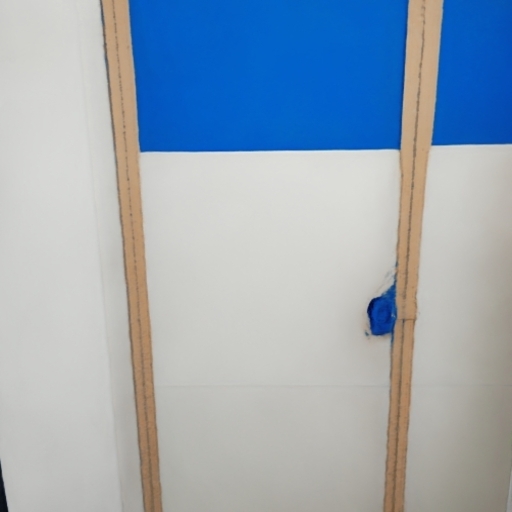}
        \hspace{0.0005\textwidth}
        \includegraphics[width=0.115\textwidth]{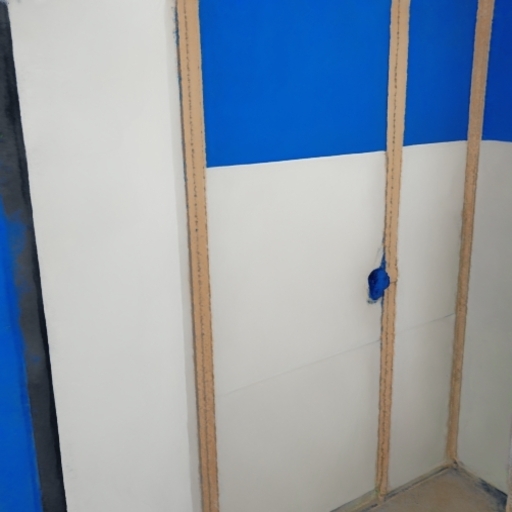}
        \hfill
        \includegraphics[width=0.115\textwidth]{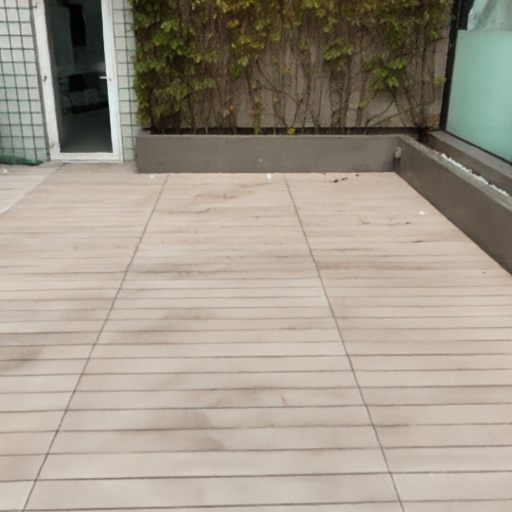}
        \hspace{0.0005\textwidth}
        \includegraphics[width=0.115\textwidth]{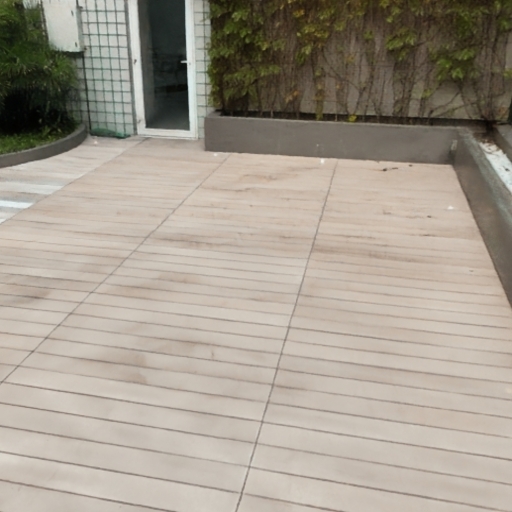}
        \hfill
        \includegraphics[width=0.115\textwidth]{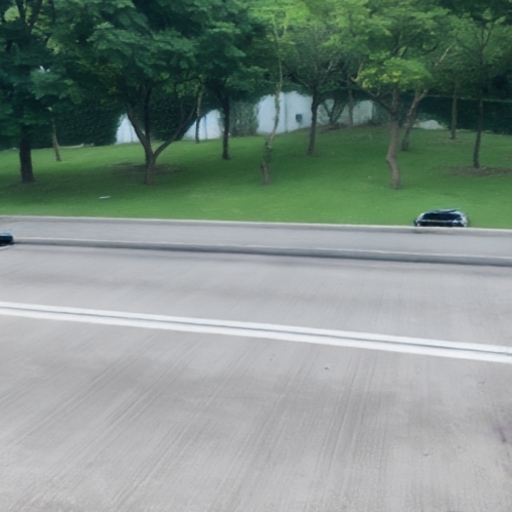}
        \hspace{0.0005\textwidth}
        \includegraphics[width=0.115\textwidth]{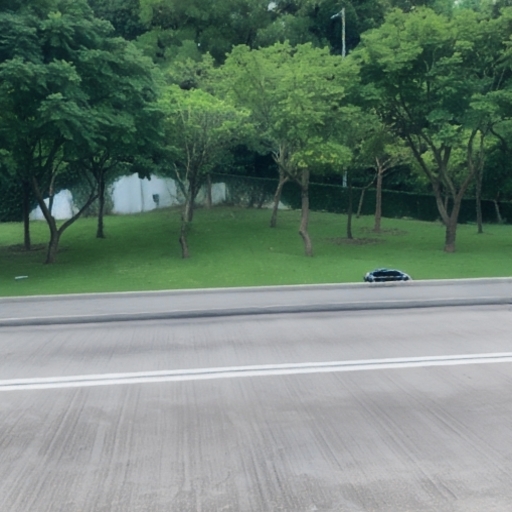}
    \end{minipage}}
    \smallskip

    \subfloat[Ours]{\begin{minipage}[c]{\textwidth}
        \includegraphics[width=0.115\textwidth]{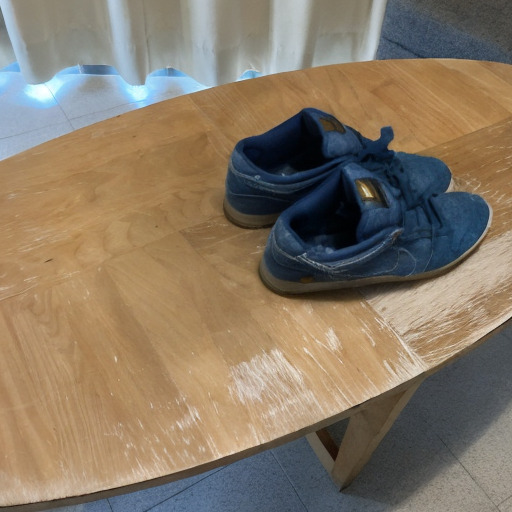}
        \hspace{0.0005\textwidth}
        \includegraphics[width=0.115\textwidth]{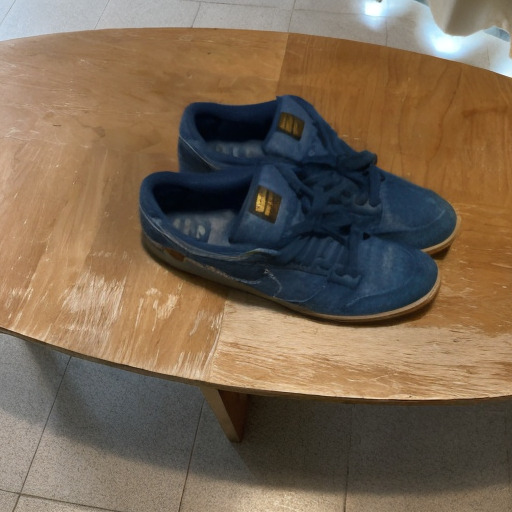}
        \hfill
        \includegraphics[width=0.115\textwidth]{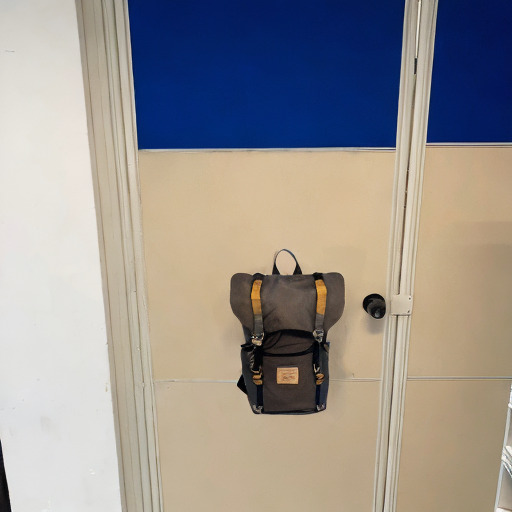}
        \hspace{0.0005\textwidth}
        \includegraphics[width=0.115\textwidth]{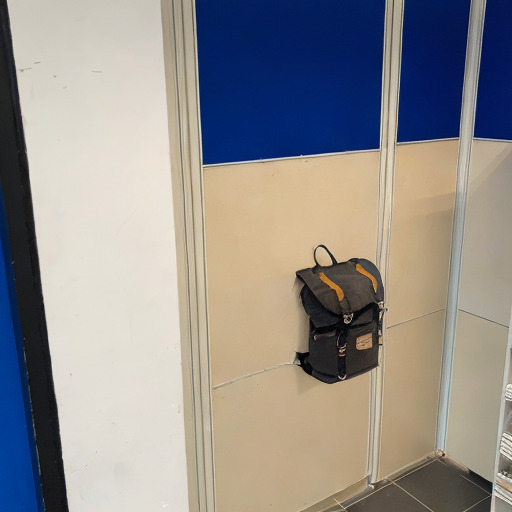}
        \hfill
        \includegraphics[width=0.115\textwidth]{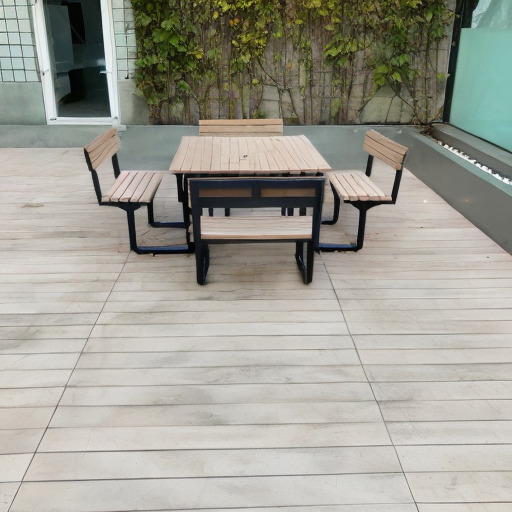}
        \hspace{0.0005\textwidth}
        \includegraphics[width=0.115\textwidth]{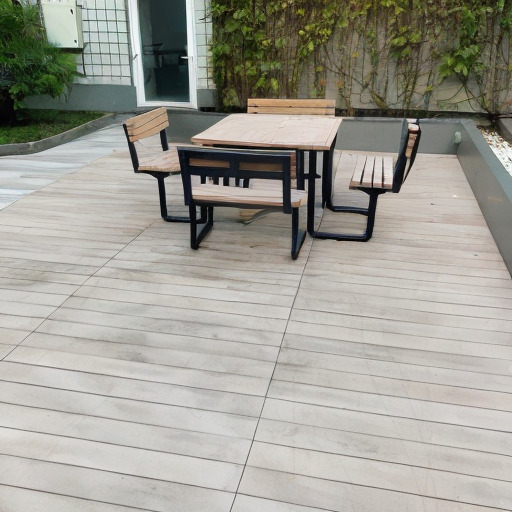}
        \hfill
        \includegraphics[width=0.115\textwidth]{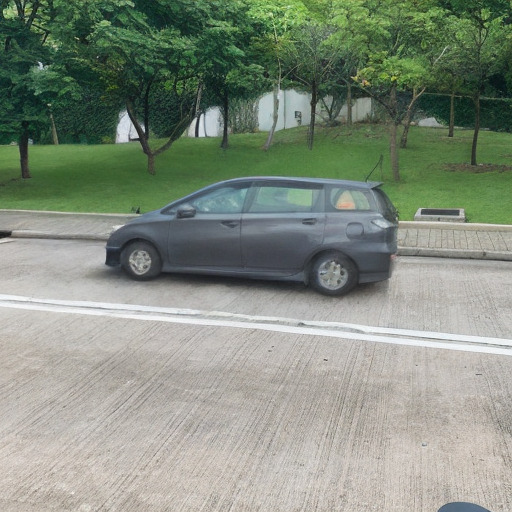}
        \hspace{0.0005\textwidth}
        \includegraphics[width=0.115\textwidth]{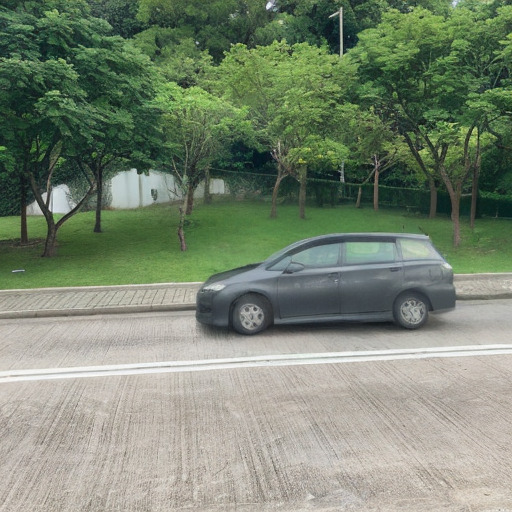}
    \end{minipage}}
    %\smallskip
    
    \caption{\textbf{Qualitative results of object insertion}. Inputs include multi-view object/background images, a 3D bounding box where the object is inserted in, and a text prompt. Note that some baselines use parts of the inputs due to the nature of their techniques.} 
    \label{fig:qualitative_baseline}
\end{figure*}

\noindent\textbf{Object insertion.} We present qualitative results of object insertion in~\cref{fig:qualitative_baseline}. As shown, our method (\cref{fig:qualitative_baseline}-f) can generate plausible contents with view consistency. Moverover, the outputs of our method also match well the text prompts and the generated objects are precisely located. 

In contrast, the traditional 3D pipeline (\cref{fig:qualitative_baseline}-b) suffers from seamed object-background boundaries and unrealistic lighting. The geometry in obscure regions, e.g., corners, is not accurately reconstructed. The image inpainting baseline (\cref{fig:qualitative_baseline}-c) produces reasonable results on individual views but generates view-inconsistency. Single-image-to-3D NeRF (\cref{fig:qualitative_baseline}-d) fails on complex scenes. A larger camera pose shift can result in background mismatch or even content collapse. Instruct-NeRF2NeRF (\cref{fig:qualitative_baseline}-e) is known to be strong at stylizing existing geometry but weak at generating new geometry. Under our setting, it performs random view-consistent editing, but fails to accomplish object insertion.

% figure, qualitative, baseline comparison

\begin{figure}
  \begin{subfigure}{0.322\columnwidth}
    \includegraphics[width=\textwidth]{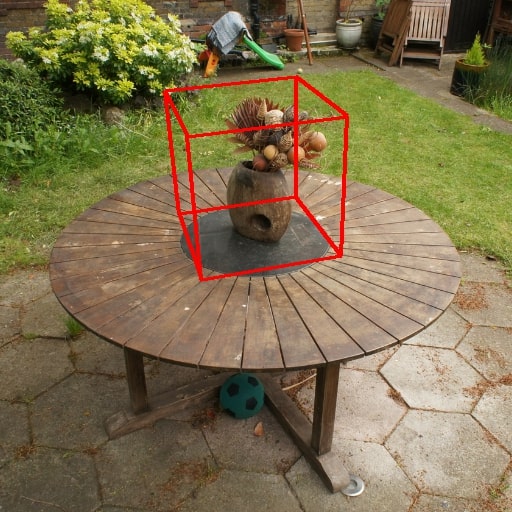}
  \end{subfigure}
  \hfill %%
  \begin{subfigure}{0.322\columnwidth}
    \includegraphics[width=\textwidth]{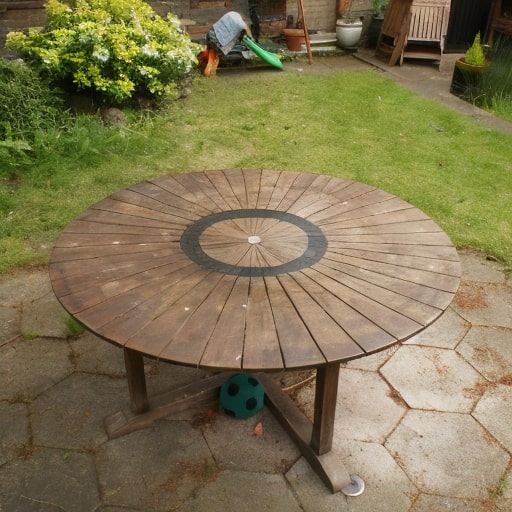}
  \end{subfigure}
  \hfill %%
  \begin{subfigure}{0.322\columnwidth}
    \includegraphics[width=\textwidth]{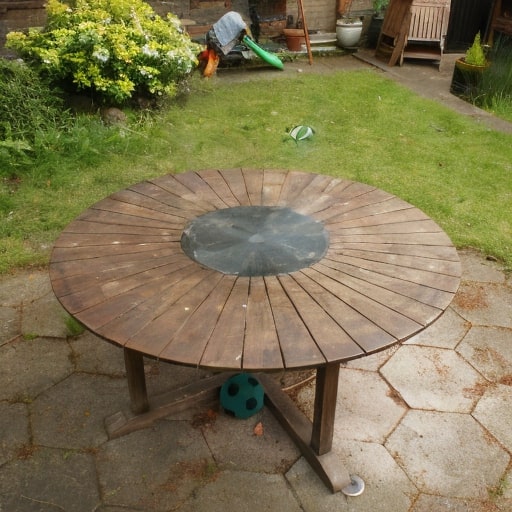}
  \end{subfigure}
  \\[0.6mm]
  \begin{subfigure}{0.322\columnwidth}
    \includegraphics[width=\textwidth]{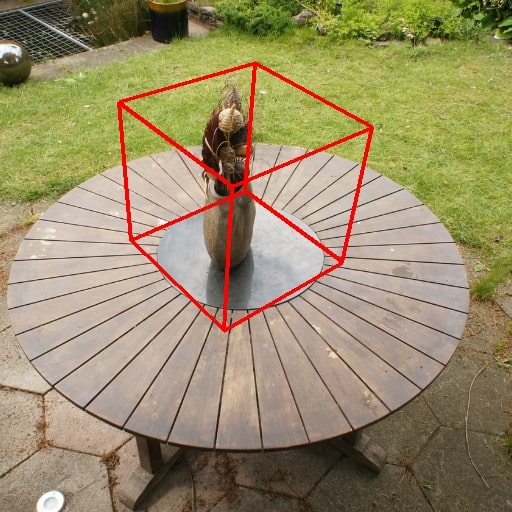}
  \end{subfigure}
  \hfill %%
  \begin{subfigure}{0.322\columnwidth}
    \includegraphics[width=\textwidth]{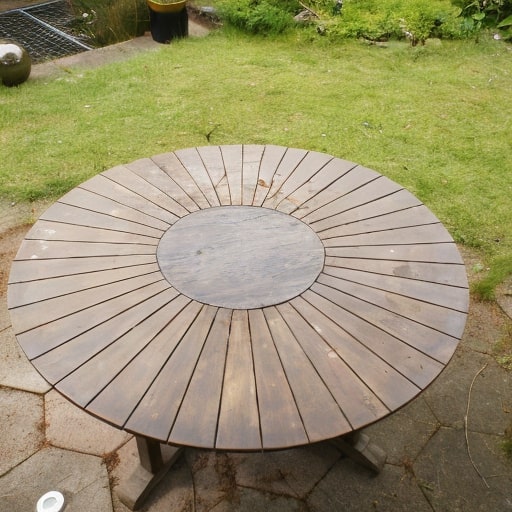}
  \end{subfigure}
  \hfill %%
  \begin{subfigure}{0.322\columnwidth}
    \includegraphics[width=\textwidth]{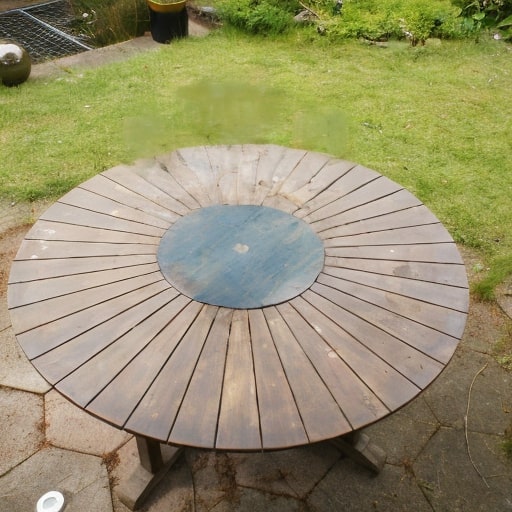}
  \end{subfigure}
  \\[2mm]
  \begin{subfigure}{0.322\columnwidth}
    \includegraphics[width=\textwidth]{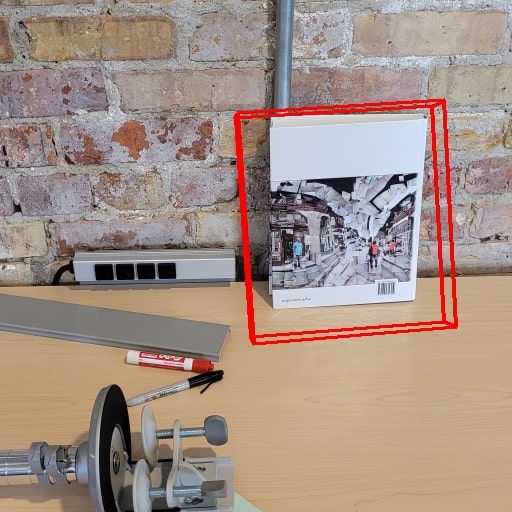}
  \end{subfigure}
  \hfill %%
  \begin{subfigure}{0.322\columnwidth}
    \includegraphics[width=\textwidth]{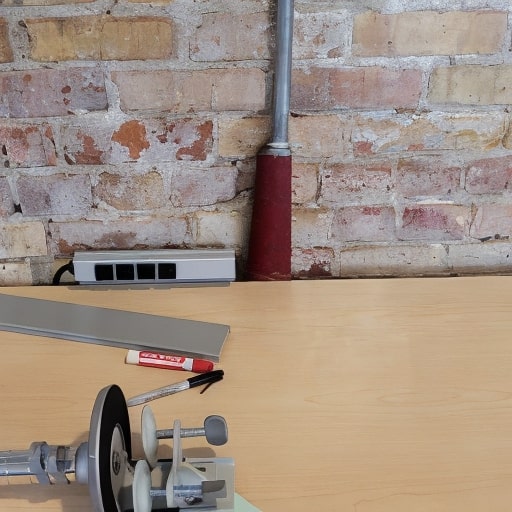}
  \end{subfigure}
  \hfill %%
  \begin{subfigure}{0.322\columnwidth}
    \includegraphics[width=\textwidth]{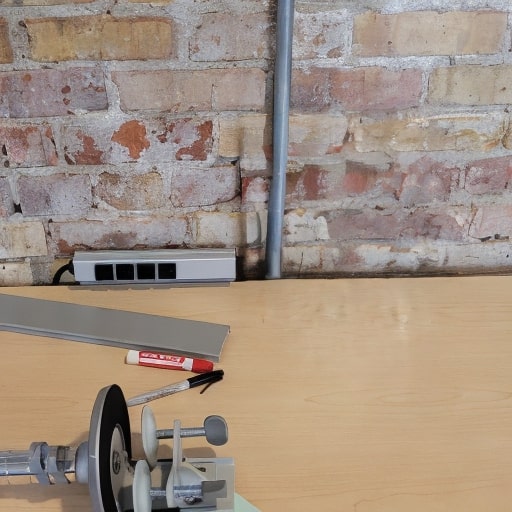}
  \end{subfigure}
  \\[0.6mm]
  \begin{subfigure}{0.322\columnwidth}
    \includegraphics[width=\textwidth]{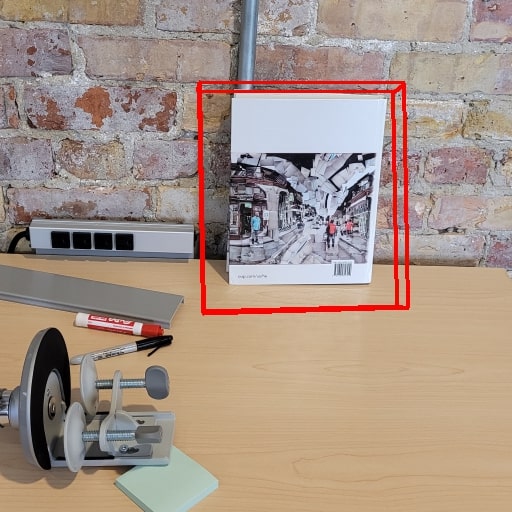}
    \caption{Original BG with object bounding box}
  \end{subfigure}
  \hfill %%
  \begin{subfigure}{0.322\columnwidth}
    \includegraphics[width=\textwidth]{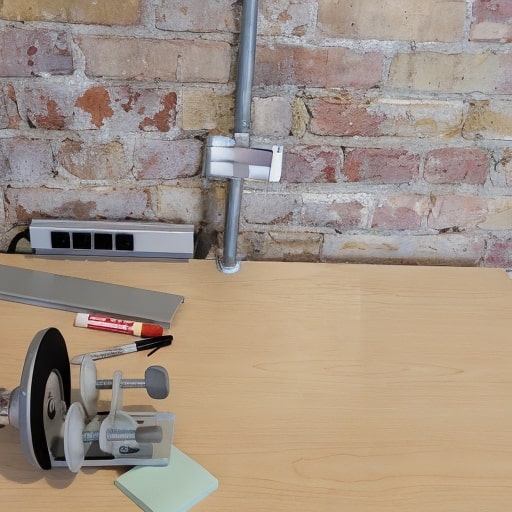}
    \caption{Pseudo ground-truth BG}
  \end{subfigure}
  \hfill %%
  \begin{subfigure}{0.322\columnwidth}
    \includegraphics[width=\textwidth]{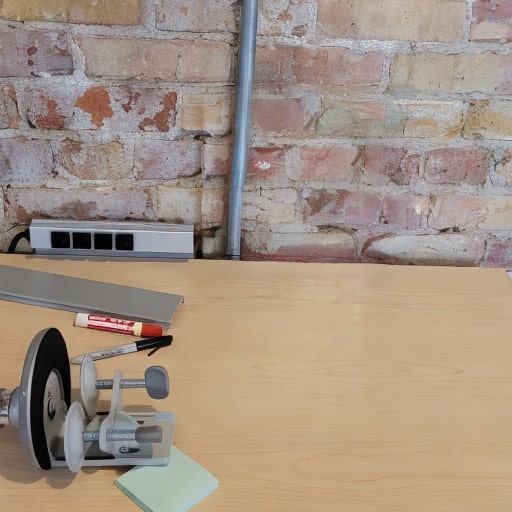}
    \caption{View-consistent editing result}
  \end{subfigure}
    
\caption{\textbf{Qualitative results of object removal} where we show the importance of pseudo ground-truth background (BG) in generating view-consistent editing results. 
%The input images with masks defined by a 3D bounding box (a) are inpainted inconsistently as pseudo background (b) for background fine-tuning. Following our pipeline, our method still converges to a consistent background across views (c).
}
\label{fig:qua_object_removal}
\end{figure}

\noindent\textbf{Object removal.}
We illustrate several results of object removal in~\cref{fig:qua_object_removal}, which shows that our method can generate view-consistent backgrounds. We also observed that without using pseudo ground-truth background, the removal can cause gradual background collapse, which we discuss further in our ablation study in~\cref{sec:ablation_study}.

% figure, qualitative, object removal

\subsection{Quantitative results}
Since there is no real-world ground-truth for scene editing, direct quantitative evaluations are not possible. Instead, we adopt the CLIP Score~\cite{hessel2021clipscore}, a well-known metric to measure how well an image correlates to a target prompt in the CLIP space. Let $E(I)$ and $E(p)$ be the CLIP embeddings of an image $I$ and a text prompt $p$. The CLIP Score is defined as,
\begin{equation}
    \mathrm{CLIPScore}(I,p) = \cos^+\left(E(I), E(p)\right)
  \label{eq:clip_score}
\end{equation}
where $\cos^+(a,b) = \max(0, \cos(a, b))$.
We average the $\mathrm{CLIPScore}(\hat{I}_b,\Tilde{p})$ for all the edited images $\hat{I}_b \in \hat{\mathbf{I}}_b$ paired with their target prompts $\Tilde{p}$, and use this average value as a performance metric for scene editing; higher score means better performance.

We also use the CLIP Directional Consistency (CLIPDC) in~\cite{haque2023instruct} to measure the editing quality and consistency across views in the CLIP space. This metric relies on an assumption that the difference in the CLIP space between a background image $I_b$ and its edited version $\hat{I}_b$ should match with the difference between the original prompt $p$ (paired with $I_b$) and the customized prompt $\Tilde{p}$ (paired with $\hat{I}_b$). Moreover, a good editing should make consistent differences between every pair of $I_b$ and $\hat{I}_b$, especially for adjacent views. We follow~\cite{haque2023instruct} to calculate the CLIPDC between two adjacent edited views $\hat{I}_b^{(i)}$ and $\hat{I}_b^{(i+1)}$ as,
\begin{equation}
\begin{split}
    & \mathrm{CLIPDC}(\hat{I}_b^{(i)},\hat{I}_b^{(i+1)}) = \\
    & \quad \cos^+\left(E(\Tilde{p}) - E(p), E(\hat{I}_b^{(i)}) - E(I_b^{(i)})\right) \times \\
    & \quad \cos^+\left(E(\hat{I}_b^{(i)}) - E(I_b^{(i)}), E(\hat{I}_b^{(i+1)}) - E(I_b^{(i+1)})\right)
    \label{eq:clip_directional_consistency}
\end{split}
\end{equation}

We measure the CLIPDC of an edited scene by averaging the $\mathrm{CLIPDC}(\hat{I}_b^{(i)},\hat{I}_b^{(i+1)})$ for all the adjacent image pairs $(\hat{I}_b^{(i)},\hat{I}_b^{(i+1)})$ in the scene. 
%We report both the CLIPScore and CLIPDC of our method and other baselines on the six scenes in our datasets in~\cref{tab:quantitative_main}.
We report both the CLIPScore and CLIPDC of our method and other baselines on 35 unique edits (cross editing on the more generalizable 5 scenes $\times$ 7 objects) in~\cref{tab:quantitative_main}. 
Experimental show that our method outperforms all the baselines on both the CLIPScore and CLIPDC metrics.

% table, quantitative, baseline comparison
\begin{table}[]
    %\small
    \centering
    \begin{tabular}{lcc}
        \toprule
        Method & CLIPScore $\uparrow$ & CLIPDC $\uparrow$ \\
        \midrule
        Traditional 3D~\cite{schoenberger2016sfm,schoenberger2016mvs} & 0.2648 & 0.1169\\ 
        Inpainting~\cite{rombach2022high} & 0.2683 & 0.1325\\ 
        Zero123~\cite{liu2023zero} & 0.2197 & 0.0361\\
        %BlendNeRF~\cite{kim20233d} & 0.xxxx & 0.xxxx\\
        Instruct-NeRF2NeRF~\cite{haque2023instruct} & 0.2347 & 0.0263\\
        \midrule
        Ours & \textbf{0.2743} & \textbf{0.1642}\\ 
        \bottomrule
    \end{tabular}
    \caption{\textbf{Quantitative results}. Higher CLIP metrics scores indicate higher image editing quality or consistency.}
    \label{tab:quantitative_main}
\end{table}

%\noindent\textbf{Run-time Statistics.}
%Object and background fine-tuning time.
%NeRF optimization time increases with a larger multi-view background dataset.

% figure, qualitative, ablation study
\begin{figure*}[t]
    %\hspace{0.6mm}sneaker on table\hspace{0.1mm}
    %\includegraphics[width=0.08\paperwidth]{images/qualitative/sneaker-table/obj/IMG_5314.png}
    %\hspace{1mm}backpack on wall\hspace{0.1mm}
    %\includegraphics[width=0.08\paperwidth]{images/qualitative/backpack-wall/obj/IMG_6483.png}
    %\hspace{1.5mm}table set in room\hspace{0.1mm}
    %\includegraphics[width=0.08\paperwidth]{images/qualitative/tableset-room/obj/IMG_6269.png}
    %\hspace{7.0mm}car on road\hspace{1.0mm}
    %\includegraphics[width=0.08\paperwidth]{images/qualitative/car-road/obj/IMG_6708.png}
   %\\[0.8mm]
    \subfloat[Fine-tuning of the diffusion model without using object images]{\begin{minipage}[c]{\textwidth}
        \includegraphics[width=\textwidth]{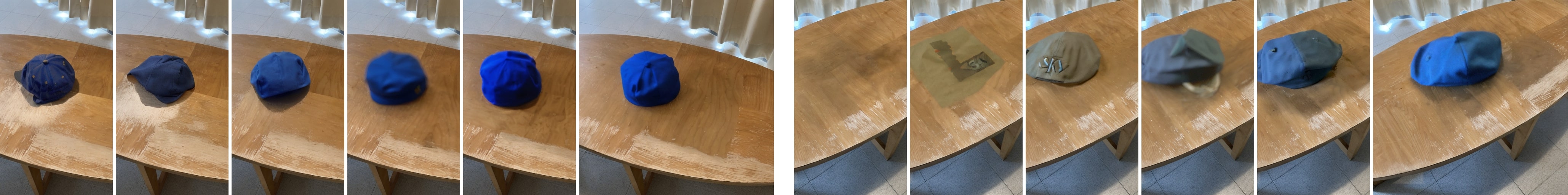}
    \end{minipage}}
    %\smallskip
    
    \subfloat[Fine-tuning of the diffusion model without using background images]{\begin{minipage}[c]{\textwidth}
        \includegraphics[width=\textwidth]{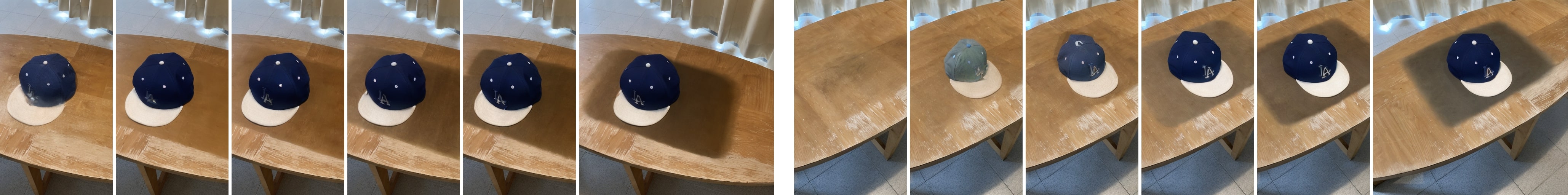}
    \end{minipage}}
    %\smallskip

    \subfloat[Dataset updates with random poses (without pose-conditioned updates)]{\begin{minipage}[c]{\textwidth}
        \includegraphics[width=\textwidth]{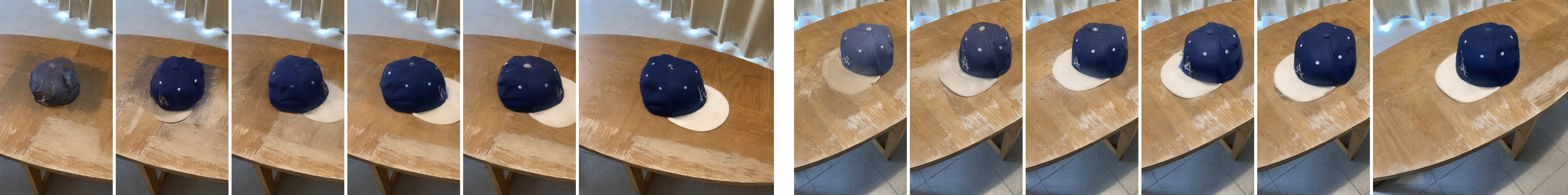}
    \end{minipage}}
    %\smallskip

    \subfloat[Without periodic dataset updates]{\begin{minipage}[c]{\textwidth}
        \includegraphics[width=\textwidth]{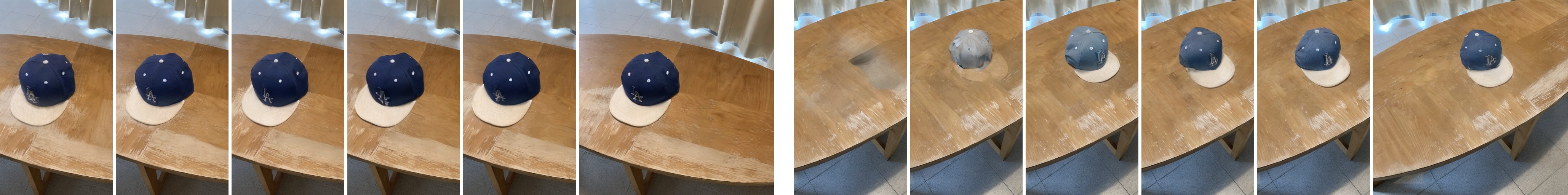}
    \end{minipage}}
    %\smallskip
    
    \subfloat[Full pipeline]{\begin{minipage}[c]{\textwidth}
        \includegraphics[width=\textwidth]{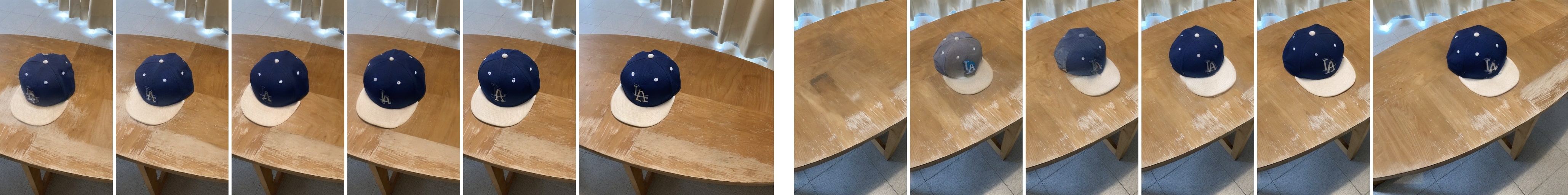}
    \end{minipage}}
    %\smallskip
    
    \caption{\textbf{Ablation study results}. Each row shows the results of a variant of our pipeline. The left and right columns include images of two different views of an edited scene. For each column, from left to right are the results of increasing training steps, where the most left image in each column is an early-stage result and the most right is the final output. The left column is a view near the starting view, which converges faster than the right column from a farther view (except for the variant in (c) where both views converge equally fast).}
    \label{fig:ablation}
\end{figure*}

\subsection{Ablation Study}
\label{sec:ablation_study}

In this ablation study, we validate the effectiveness of technical components in our method. We suggest readers observe the results in~\cref{fig:ablation}, which shows two views of an edited scene. We provide detailed analyses on these results below.

\noindent\textbf{Diffusion model fine-tuning.}
Recall that we fine-tune the diffusion model $D_\theta$ on both object and background images (in~\cref{sec:fine-tuning}). Here we prove that such a fine-tuning is necessary. Fine-tuning on object images makes the diffusion model aware of the same object during the dataset updates (in~\cref{sec:near_to_far_NeRF}). We verify this in~\cref{fig:ablation}-a, where we skip the fine-tuning of $D_\theta$ on object images. As shown, without using object images, the diffusion model cannot generate the same object across views.

Likewise, fine-tuning the diffusion model with background images helps to preserve the background at inpainted borders (areas in between mask boundary and inner-mask object boundary). As shown in~\cref{fig:ablation}-b, without fine-tuning on background images, the model cannot inpaint the background within the mask region properly. The inpainted background gets darker at latter NeRF updating steps, and finally ends up with a noticeable error. We hypothesize this collapse that the model, without being fine-tuned on relevant background images, applies the bias from pre-learned knowledge which does not align with the surrounding background. 

%Likewise, fine-tuning the diffusion model with background images helps the model not only preserve the background outside of inpainted regions (masks) but also create seamless perception at inpainted borders (between outer-mask background and inner-mask background). As shown in~\cref{fig:ablation}-b, without fine-tuning on background images, the model cannot inpaint the background within the mask region properly. The inpainted background gets darker at latter training step, and finally ends up with a noticeable error. We hypothesize this collapse that the model, without being fine-tuned on relevant background images, applies the bias from pre-learned knowledge which does not align with the surrounding background. 

%Fine-tuning on background images helps the diffusion model revise the wrong pixels to the learned background. Note that a high noise strength setting partially solve this issue but causes another consistency problem, which we will discuss in a later subsection in this ablation study.

\noindent\textbf{Pose-conditioned dataset updates.}
To prove the effectiveness of the pose-conditioned dataset update strategy, we experiment with another scheme in which the dataset used to refine the NeRF is updated with random views. We present the results of this experiment in~\cref{fig:ablation}-c. We observe that, with view-random updates, objects in different views can converge into inconsistent poses (e.g., the hats in the left and right view are generated in different poses). 

%This inconsistency 

The above inconsistency commonly happens with many text-to-3D NeRF generation or editing methods. We hypothesize this phenomenon as follows. Initially the diffusion model can generate the object in inconsistent poses across views due to little 3D-aware hint available. This inconsistency is continually introduced to the NeRF updating and then, in return, passed partially as input to the diffusion model (as in~\cref{eq:near_view_generation}), making further pose divergence eventually. Without considering nearby-views, the diffusion model cannot fix this inconsistency as objects generated on individual views match well their given text prompts. 

%Since little hint is available at the start of the fine-tuning, the diffusion model may generate objects with different poses across view points. This inconsistency is continually introduced to the NeRF updating and then, in return, passed as input to the diffusion model to further diverge the view consistency. The diffusion model has no intent to fix such an inconsistency as objects generated on individual views match well their given text prompts. This problem is also known as the Janus Problem, which commonly exists in many text-to-3D NeRF generation or editing methods.

%and optimize all views in the NeRF training dataset randomly and uniformly. Without the pose-conditioned strategy, objects in different views may converge to inconsistent poses. As little hint is available at the start of training, the diffusion model may generate objects with different poses across view points. This inconsistency is continually introduced into NeRF training and then in return, be passed as diffusion model input to further strengthen the view difference. The diffusion model has no intent to fix such an inconsistency as these objects are individually good enough to match the text prompt and the fine-tuned object target. This problem is also known as the Janus Problem, which commonly exists in many text-to-3D NeRF generation or editing methods.

Our proposed pose-conditioned dataset update strategy simulates the nature of NeRF construction, in which a view rendered closed to already-used views should contain similar but slightly blurry and distorted object content. Pose-conditioned view arrangement thus gives the diffusion model enough hints to generate objects with view-consistent appearance and poses, specified in nearby views. View-consistency is thus achieved progressively in the training dataset and is integrated into the constructed NeRF (see~\cref{fig:ablation}-e). %We provide more analysis of this convergence in our supplementary.

%Due to the nature of the NeRF parameters, a view rendered closed to the already-used views should contain similar but slightly blurry and distorted object content. It gives the diffusion model enough hints to generate objects of similar and more correct pose in these nearby views. The consistent information thus progressively expands to the entire dataset and is fused as 3D in NeRF.

\noindent\textbf{Periodic dataset updates.}
We observe that periodically updating of training views during the NeRF updating is important to achieve high-quality renderings. We verify this by keeping all the training views from the pose-conditioned dataset updates fixed during the NeRF updating. We found that although objects are rendered at fairly accurate orientations, their texture and geometry still suffer from inconsistencies (see~\cref{fig:ablation}-d). Having periodically updates on the training data fixes these minor defects and facilitates a more consistent convergence of the NeRF updating (see~\cref{fig:ablation}-e).

%when we keep the training views from the pose-conditioned dataset updates fixed in the entire NeRF updating process, we found that although objects in the rendering have fairly accurate positions and orientations, their texture and geometry suffer from inconsistencies. 

%\subsection{User Study} 
%Image quality: coherency of geometry, lighting. 

%\subsection{Controllability}

\section{Conclusion}

%iterative dataset updates that leverages a text-to-image diffusion model to fuse object into background images and a pose-conditioned dataset updates strategy to stabilize the NeRF training when the object manipulation is progressively introduced into the training process. 

We propose a new language-driven method for object manipulation in NeRFs. Our method is built upon a novel idea of joint 2D-3D interaction, keeping both 2D image synthesis and 3D NeRF reconstruction in a loop. This idea is enabled by an advanced text-to-image diffusion technique that generates object-blended background images, and a novel pose-conditioned dataset update strategy that learns a NeRF from the multi-view images in a progressive manner. 
%Our method shows promising results on a diverse set of test scenes, and in two tasks: object insertion and object removal. 

Our method is not without limitations. Since our 2D views are synthesized by a diffusion model, we may share the flickering problem with diffusion-based video editing methods~\cite{wu2022tune,kim2023diffusion,qi2023fatezero}. We leave this for future work, which can potentially be addressed by robust video translation methods~\cite{yang2023rerender}.
%Second, extending our method to support more tasks, e.g., object translation and rotation, would result in more comprehensive editing capabilities. 
It is also of great interest to postulate a theoretical foundation for pose-conditioned dataset update to better understand the convergence of NeRF training in scene editing. 

%------------------------------------------------------------------------

\noindent\textbf{Acknowledgment.} This work was partially supported by a grant from the RGC of HKSAR, China (Project No. HKUST 16202323) and an internal grant from HKUST (R9429).

%%%%%%%%% REFERENCES
{\small
\bibliographystyle{ieee_fullname}
\bibliography{egbib}
}

\end{document}